\pdfoutput=1
\documentclass[10pt,twocolumn,letterpaper]{article}
\usepackage[pagenumbers]{iccv} %

\usepackage{multirow}
\usepackage{makecell}
\usepackage[dvipsnames]{xcolor}
\usepackage{comment}
\usepackage{pifont}%
\usepackage{colortbl}
\usepackage{tabularx}
\usepackage{adjustbox}
\usepackage[frozencache,cachedir=.]{minted} %
\makeatletter
\newcommand*{\@rowstyle}{}

\newcommand*{\rowstyle}[1]{%
  \gdef\@rowstyle{#1}%
  \@rowstyle\ignorespaces%
}

\newcolumntype{=}{%
  >{\gdef\@rowstyle{}}%
}

\newcolumntype{+}{%
  >{\@rowstyle}%
}
 \makeatother

\newcommand\imgclip[2]{\adjincludegraphics[Clip={#1\width} {#1\height} {#1\width} {#1\height}]{#2}}

\newcommand{\mypara}[1]{\noindent\textbf{#1}}
\newcommand{\cmark}{\textcolor{tblgreen}{\ding{51}}}%
\newcommand{\xmark}{\textcolor{red}{\ding{55}}}%

\definecolor{tblblue}{RGB}{31, 119, 180}
\definecolor{tblorange}{RGB}{255, 127, 14}
\definecolor{tblgreen}{RGB}{44, 160, 44}
\definecolor{tblred}{RGB}{214, 39, 40}

\definecolor{babyblue}{RGB}{208, 254, 254}
\definecolor{offwhite}{RGB}{255, 255, 228}

\newcommand{\R}[1]{{%
    \textbf{%
        \ifstrequal{#1}{1}{\textcolor{tblred}{R#1}}{%
        \ifstrequal{#1}{2}{\textcolor{tblblue}{R#1}}{%
        \ifstrequal{#1}{3}{\textcolor{tblorange}{R#1}}{%
                           \textcolor{tblgreen}{R#1}
        }}}%
    }%
}}

\definecolor{iccvblue}{rgb}{0.21,0.49,0.74}
\usepackage[pagebackref,breaklinks,colorlinks,allcolors=iccvblue]{hyperref}

\def\paperID{10367} %
\def\confName{ICCV}
\def\confYear{2025}

\title{Diorama: Unleashing Zero-shot Single-view 3D Indoor Scene Modeling}
\author{
Qirui Wu$^{1}$ \quad 
Denys Iliash$^{1}$ \quad 
Daniel Ritchie$^{2}$ \quad
Manolis Savva$^{1}$ \quad 
Angel X. Chang$^{1,3}$\\
$^{1}$Simon Fraser University \quad
$^{2}$Brown University \quad
$^{3}$Alberta Machine Intelligence Institute (Amii)\\
{\tt\small \url{https://3dlg-hcvc.github.io/diorama/}}
}

\begin{document}

\newcommand{\teaserfigure}{
\vspace{-2.75em}
\begin{center}
\captionsetup{type=figure}
\includegraphics[width=\textwidth]{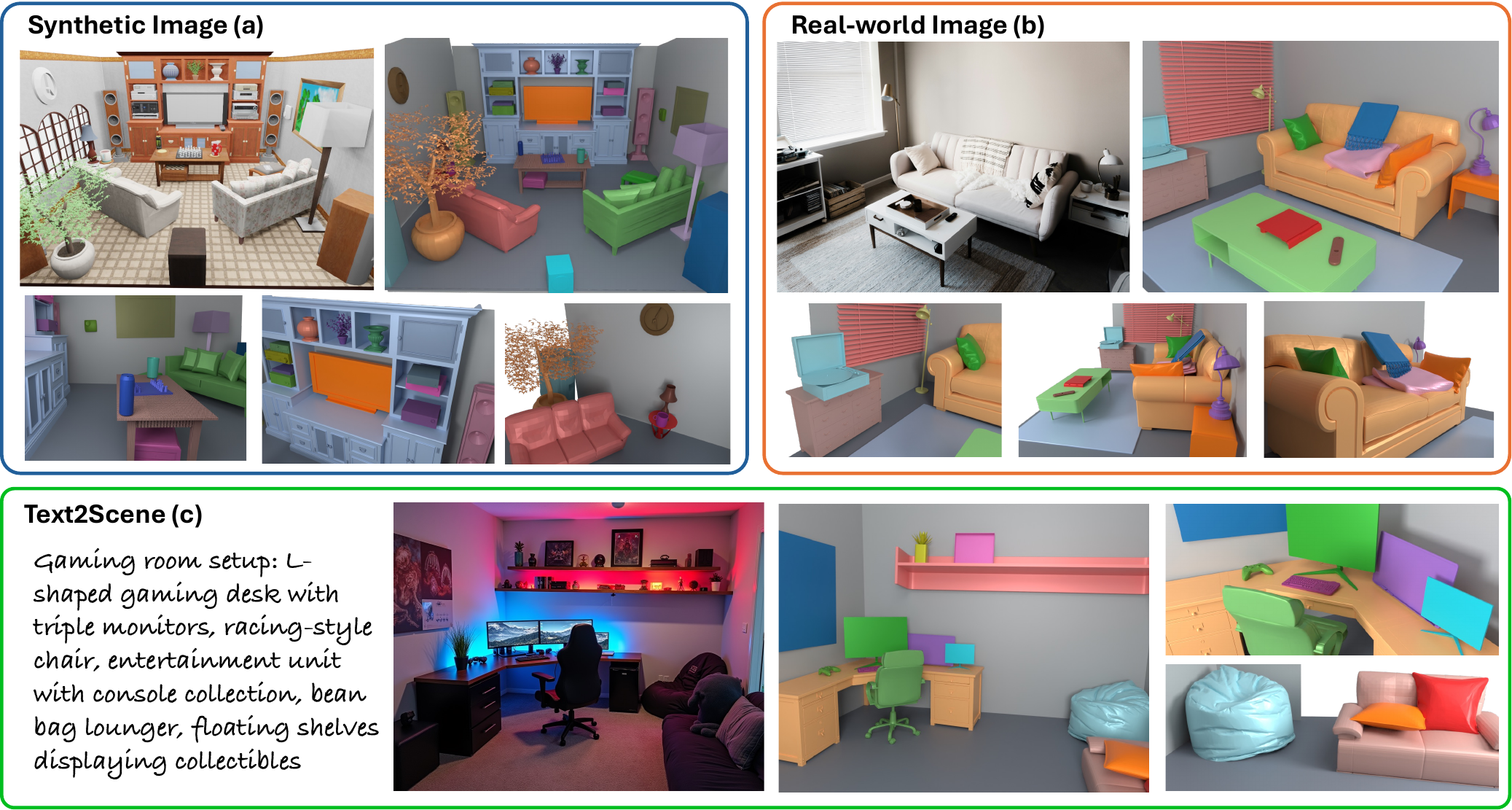}
\captionof{figure}{
We propose Diorama: a system for zero-shot single-view 3D scene modeling.
Our system produces a holistic 3D scene model given a single input image, representing both architectural elements and objects in cluttered indoor scenes.
We showcase three examples of our system where the input image is shown in the top left in each case.
Other images are renderings of the output 3D scene from different camera viewpoints.
\textbf{(a)} 3D scene model from a synthetic image input.
\textbf{(b)} 3D scene model from a real-world internet image input.
\textbf{(c)} Text-to-scene pipeline where the input image on the left is generated from a text prompt.
}
\vspace{-6pt}
\label{fig:teaser}
\end{center}

\vspace{0.5em}
}

\twocolumn[{
\vspace*{-0.5cm}
\maketitle

\vspace{-2.75em}
\begin{center}
\captionsetup{type=figure}
\includegraphics[width=\textwidth]{fig/images/teaser.pdf}
\captionof{figure}{
We propose Diorama: a system for zero-shot single-view 3D scene modeling.
Our system produces a holistic 3D scene model given a single input image, representing both architectural elements and objects in cluttered indoor scenes.
We showcase three examples of our system where the input image is shown in the top left in each case.
Other images are renderings of the output 3D scene from different camera viewpoints.
\textbf{(a)} 3D scene model from a synthetic image input.
\textbf{(b)} 3D scene model from a real-world internet image input.
\textbf{(c)} Text-to-scene pipeline where the input image on the left is generated from a text prompt.
}
\vspace{-6pt}
\label{fig:teaser}
\end{center}

\vspace{0.5em}

}]

\maketitle

\begin{abstract}

Reconstructing structured 3D scenes from RGB images using CAD objects unlocks efficient and compact scene representations that maintain compositionality and interactability.
Existing works propose training-heavy methods relying on either expensive yet inaccurate real-world annotations or controllable yet monotonous synthetic data that do not generalize well to unseen objects or domains.
We present Diorama, the first zero-shot open-world system that holistically models 3D scenes from single-view RGB observations without requiring end-to-end training or human annotations.
We show the feasibility of our approach by decomposing the problem into subtasks and introduce better solutions to each: architecture reconstruction, 3D shape retrieval, object pose estimation, and scene layout optimization.
We evaluate our system on both synthetic and real-world data to show we significantly outperform baselines from prior work.
We also demonstrate generalization to real-world internet images and the text-to-scene task.

\end{abstract}
    
\section{Introduction}

Modeling a 3D scene from a single image input is a cornerstone task in computer vision that is critical in real-world applications such as autonomous driving, augmented/virtual reality, and robotic perception.
Recent works tackling 3D perception as a reconstruction problem tend to produce imperfect surface-only 3D meshes that are incompatible with modern graphics pipelines or simulation platforms~\cite{nie2020total3dunderstanding, liu2022towards, chen2024single}. %
An alternative strategy leverages a database of CAD models to model a 3D scene in an analysis-by-synthesis manner~\cite{kuo2020mask2cad, gumeli2022roca, gao2023diffcad} that enables convenient scene creation and editing.
Such representation, even with mismatched exterior geometry and texture, can be easily applied for robot learning~\cite{dai2024acdc, zook2024grs}, leveraging human-labeled annotations, including physical properties, articulation parameters and complete interior geometry.

Creating 3D scenes typically involves intensive manual work by 3D designers.
The resulting 3D scenes are oversimplified~\cite{fu20213d} or of limited accessibility to the public with only images available~\cite{zheng2020structured3d, roberts2021hypersim}.
More recent works address the synthetic data domain gap by aligning CAD objects to real-world RGBD scans~\cite{avetisyan2019scan2cad, szot2021habitat, liu2023lasa}, RGB videos~\cite{maninis2023cad} or panoramas~\cite{wu2024r3ds}. However, they either ignore architecture or do not reach the scale appropriate for training.

Recent studies show that foundation models exhibit 3D awareness~\cite{el2024probing, yue2025improving, wang2024lift3d}.
This paper is motivated by the question \textit{``Is holistic 3D scene modeling from a single-view real-world image possible using foundation models?''}
To answer this question we present Diorama: a modular zero-shot open-world system that models synthetic 3D scenes given an image and requires no end-to-end training.

Our system consists of two major components:
1) \textit{Open-world perception for holistic scene understanding} of the input image, including object recognition and localization, depth and normal estimation, architecture reconstruction and scene graph generation.
2) \textit{CAD-based scene modeling} for assembly of a clean and compact 3D scene through CAD model retrieval, 9-DoF pose estimation, and semantic-aware scene layout optimization.
Different from other zero-shot modular scene generation systems focusing on texture stylization~\cite{huang2023aladdin}, object reconstruction~\cite{zhou2024zero} and language description~\cite{fu2025anyhome, aguina2024open}, our work focuses on open-world holistic perception of objects, architecture, and spatial relationships.
There are also works proposing to create synthetic scenes from images for robotic manipulation. However, current efforts~\cite{li2024evaluating, dai2024acdc, agarwal2024scenecomplete, zook2024grs, chen2024urdformer} either use strong construction heuristics or only handle tabletop setups. We advocate our system as a more general approach to output cluttered compositional scenes that can be rearranged and edited.
In summary, our contributions are:
\begin{itemize}
    \item We introduce Diorama, the first zero-shot open-world system for CAD retrieval-based 3D scene modeling from a monocular observation.
    \item We ensure flexibility and generalization to real-world data with a modular design, proposing better and more robust solutions to planar architecture reconstruction, 3D shape retrieval, zero-shot 9D pose estimation, and stage-wise scene layout optimization.
    \item We benchmark on synthetic and real-world data, and demonstrate the usability of our approach on real-world internet images and text-to-scene applications.
\end{itemize}

\begin{figure*}[t]
\centering
\includegraphics[width=\linewidth]{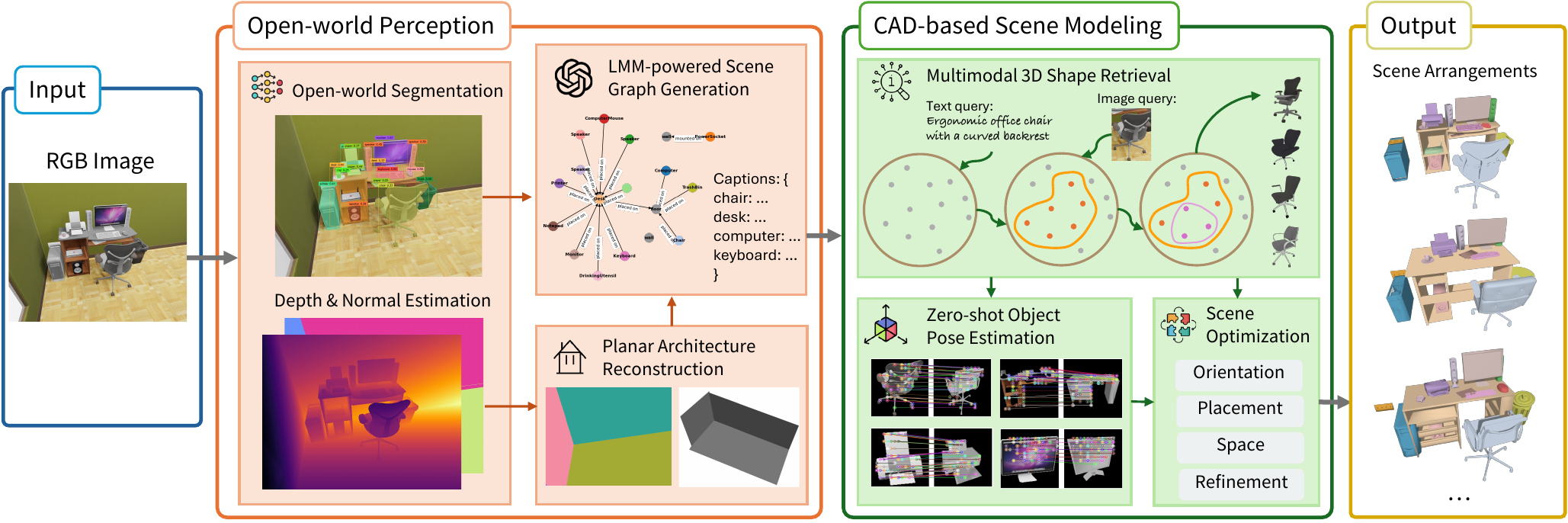}
\caption{
Illustration of the Diorama pipeline.
The input image is processed in the open-world perception component (in orange box) through object instance segmentation, depth and normal estimation, architecture reconstruction and LLM-powered scene graph generation.
The CAD-based scene modeling component (green box) then assembles a compositional 3D scene by retrieving and posing objects from a database and optimizing the overall scene layout.
Multiple plausible scene arrangement hypotheses are produced as outputs.
} 
\vspace{-8pt}
\label{fig:pipeline}
\end{figure*}

\section{Related work}

We summarize prior work on open-world 3D perception, single-view scene modeling and reconstruction, and generative models for scene synthesis.

\mypara{Open-world 3D perception.}
Visual foundation models enabled various aspects of open-world 3D perception, including 3D shape understanding~\cite{zhu2023pointclip, xue2024ulip, liu2024openshape, lee2024duoduo}, part-level 3D shape recognition~\cite{liu2023partslip, abdelreheem2023satr}, 3D semantic segmentation~\cite{peng2023openscene, zhang2023clip, chen2023clip2scene, ding2023pla, jatavallabhula2023conceptfusion}, 3D instance segmentation~\cite{takmaz2023openmask3d, nguyen2024open3dis, huang2023openins3d}, 3D object detection~\cite{lu2023open, yang2024imov3d}, neural field rendering~\cite{kerr2023lerf, qin2024langsplat} and language grounding~\cite{jia2025sceneverse, delitzas2024scenefun3d}. Most works obtain semantic-rich representation for open-world language query by directly leveraging CLIP feature or learning to align with CLIP feature.
Some works~\cite{hong20233d, fu2024scene} propose to learn tokenized 3D feature along with LLMs.
In contrast, our work directly models a complex, cluttered 3D scene from a single-view image and unlike prior work does not rely on any 3D modality inputs such as point clouds.

\mypara{Single-view scene modeling.}
There is a long history of work on scene modeling through CAD object retrieval and alignment~\cite{izadinia2017im2cad, huang2018holistic, kuo2020mask2cad, kuo2021patch2cad, gumeli2022roca, sparc, gao2023diffcad, maninis2022vid2cad, langer2024fastcad}.
IM2CAD~\cite{izadinia2017im2cad} breaks down the task into separate modules: room layout estimation, object recognition and scene configuration. However, IM2CAD cannot easily generalize to unseen objects and oversimplifies complex architecture and arrangements in real-world images.
More recent works explore end-to-end trained methods~\cite{kuo2020mask2cad, kuo2021patch2cad, gumeli2022roca, sparc}.
Mask2CAD~\cite{kuo2020mask2cad} jointly learns object detection, 3D shape embedding and pose regression.
Patch2CAD~\cite{kuo2021patch2cad} introduces a more robust patch-based joint embedding space for 3D shape retrieval.
ROCA~\cite{gumeli2022roca} leverages differentiable Procrustes optimization for pose estimation, and SPARC~\cite{sparc} employs an iterative ``render-and-compare'' strategy.
DiffCAD~\cite{gao2023diffcad} achieves state-of-the-art performance with a combination of probabilistic diffusion models trained in a weakly-supervised manner on synthetic data. 
However, all existing methods do not model architecture and require significant amounts of annotated data for training.
Our system is training-free, requires no 3D supervision, and leverages large foundation models pretrained on large corpus data such that it generalizes to real-world images.

\mypara{Single-view scene reconstruction.}
There is much work on single-view scene reconstruction~\cite{irshad2022centersnap, nie2020total3dunderstanding, liu2022towards, chen2024single, zhang2021deeppanocontext, dong2024panocontext}.
Due to the limited availability of real-world data, prior works train on synthetic 3D shapes. This typically leads to limitations such as oversmoothed and incomplete meshes, discarding of fine-grained scene structure, and real to synthetic domain gaps.
Recent work implements modular zero-shot scene reconstruction using inpainting and image-conditioned 3D shape generation for each object~\cite{zhou2024zero}.
In contrast, we produce a holistic scene model including both architecture and objects, and fine-grained object relations in highly cluttered scenes not handled by this prior work.

\mypara{Generative scene synthesis.}
Another line of work to synthetic 3D scene modeling is through learning scene configuration priors from large-scale data~\cite{fu20213d}. Various neural network-based approaches have explored capturing scene layout priors using CNNs~\cite{wang2018deep, ritchie2019fast}, relation graphs~\cite{wang2019planit}, transformers~\cite{paschalidou2021atiss, wang2021sceneformer}, and diffusion models~\cite{wei2023lego, tang2024diffuscene, lin2024instructscene}.
These approaches are restricted by training data quality and scale, and generally only produce furniture layouts on a horizontal plane.
Recent works~\cite{fu2025anyhome, tam2024scenemotifcoder, gao2024graphdreamer, yang2024holodeck, aguina2024open} propose to generate layout specifications using LLMs to mitigate the training data sparsity issues.
However, these approaches struggle to accurately position objects as LLMs do not encode fine-grained 3D spatial knowledge.
In contrast, our system directly perceives images and converts them to plausible 3D layouts respecting the provided input.

\section{Approach}

We present Diorama, a zero-shot open-world system that reconstructs synthetic 3D scenes from monocular image inputs, without end-to-end training as required by prior work~\cite{gumeli2022roca,sparc,gao2023diffcad}.
Our system has two major components:
\textbf{1) Open-world perception} for holistic scene understanding of the input image, including object recognition and localization, depth and normal estimation, architecture reconstruction and scene graph generation (\cref{fig:pipeline}, orange box).
\textbf{2) CAD-based scene modeling} for assembly of a clean and compact 3D scene representation through CAD model retrieval, 9-DoF pose estimation, and semantic-aware scene layout optimization (\cref{fig:pipeline}, green box).
Our system design follows the principles of open vocabulary, category agnosticism, and robustness to non-exact object matches.
The modeled scenes benefit from certain designs of Diorama, including scene graph for maintaining spatial relationships among objects, shape retrieval for multiple semantically similar arrangements, planar architecture for physically plausible supporting objects, and layout optimization for refining objects poses based on spatial relationships.
We describe each module in the following sections, with additional details in the supplement. 

\subsection{Open-world Perception}

\mypara{Holistic scene parsing.}
Given an image $\mathbf{I}$, we recognize all objects in the image as a list of open-vocabulary categories $\mathbf{C}$.
We combine an open-vocabulary detector OWLv2~\cite{minderer2023scaling} and Segment Anything~\cite{kirillov2023segany} to localize object instance as bounding boxes $\mathbf{B}$ and masks $\mathbf{M}$.
Thus, each localized object can be represented by its category, bounding box and instance mask, $\mathcal{O}=\{o_i = \{c_i, b_i, m_i\}\}_{i=1}^N$.
We use Metric3DV2~\cite{hu2024metric3d} to estimate metric depth $\mathbf{D}$ and normal maps $\mathbf{N}$.
We then back project the estimated depth to a point cloud $\mathbf{P}$.
For each object $o_i$, we extract an instance point cloud $p_i$ from $\mathbf{P}$ using the bounding box $b_i$ and mask $m_i$.

\mypara{LMM-powered scene graph generation.} 
To model complex real-world scenes, it is essential to understand the spatial relationships between objects.
In particular, it is key to model the support hierarchy, which describes a complete static scene arrangement from distinct objects.
We follow SoM~\cite{yang2023setofmark} to leverage the visual grounding capability of Large Multimodal Model (GPT-4o). See the prompt example and details in the supplement.
The output scene graph is formalized as $G=\langle V, E\rangle$, where each vertex $v_i\in V$ represents an object instance $o_i$ in the image along with its language caption and each edge $e=\langle v_i, v_j \rangle \in E$ indicates $o_i$ is supported by $o_j$.

\begin{figure}[t]
\centering
\includegraphics[width=\linewidth]{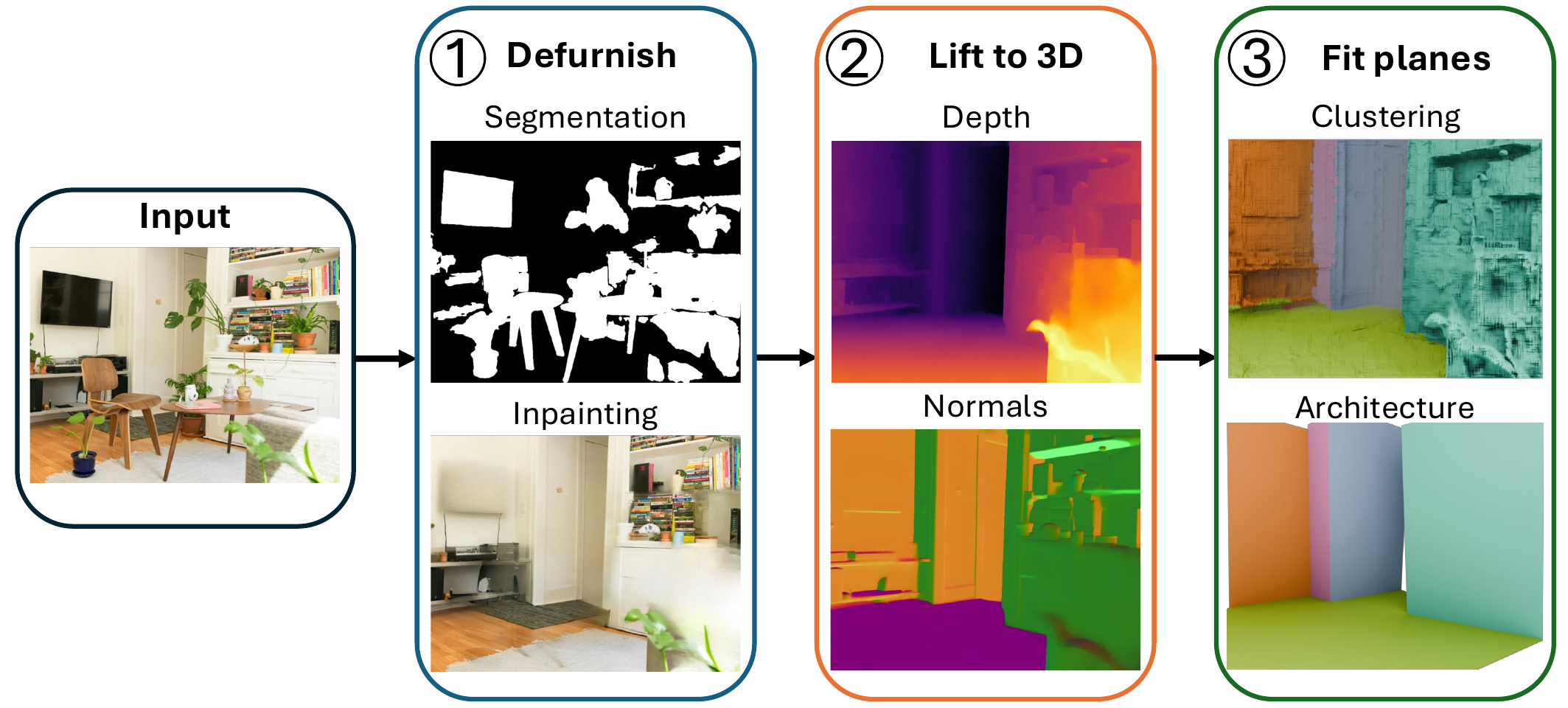}
\vspace{-6pt}
\caption{
Our PlainRecon architecture reconstruction approach.
Objects are first segmented. The object masks are inpainted, and depth and normals are estimated.
Then, normal-based clustering on the point clouds is used to produce the 3D architecture.
} 
\label{fig:arch-recon}
\end{figure}

\mypara{Planar architecture reconstruction.}
Reconstructing the surrounding architecture enables better physical plausibility of the scene arrangement, by ensuring objects are supported on surfaces and are not floating in the air.
We represent architecture as 3D planes following~\cite{stekovic2020general, rozumnyi2023estimating} to better capture general complex room layout in the real world.
We introduce \textit{PlainRecon}, a simple yet effective architecture reconstruction pipeline leveraging inpainting~\cite{yu2023inpaint}, monocular depth and normal estimation~\cite{hu2024metric3d} that handles both synthetic and real-world images.
PlainRecon consists of three steps (see \cref{fig:arch-recon}):
1) segment objects and inpaint to produce an ``empty room'' showing the complete architecture;
2) extract a point cloud and normals for each architecture element using depth estimation; and
3) fit 3D planes to architecture elements using normal-based clustering.
Given planar segments obtained from normals-based clustering, we fit a plane equation to each point cloud of an architectural element and compute bounds in each plane.
See the supplement for more technical details.

\begin{figure}[t]
\centering
\includegraphics[trim={0 8px 0 0},clip,width=\linewidth]{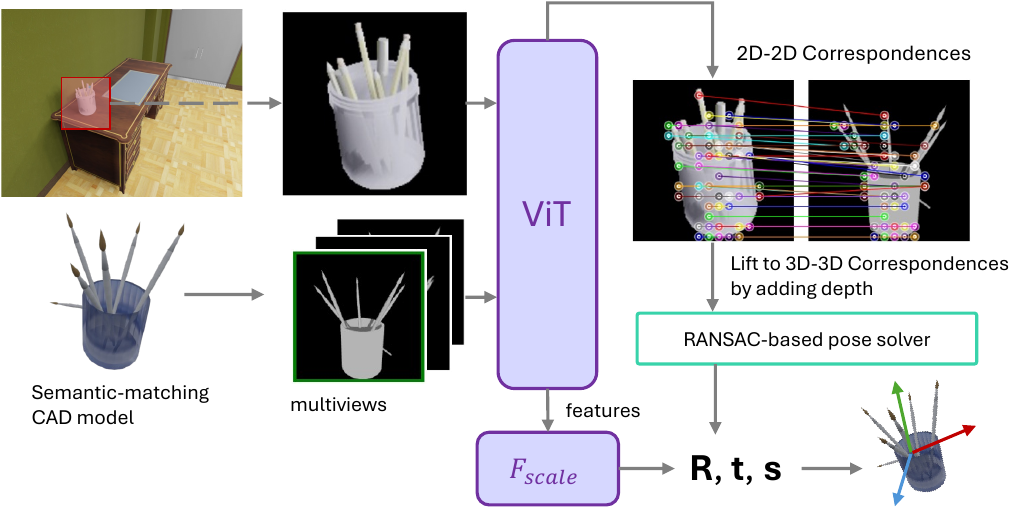}
\vspace{-12pt}
\caption{
Our zero-shot object pose estimation approach.
We leverage vision transformer features to establish 2D and 3D correspondences and estimate 9-DoF poses for each object.
} 
\label{fig:pose-estimation}
\end{figure}

\subsection{CAD-based scene modeling}

\mypara{Multimodal shape retrieval.}
We retrieve CAD objects from a combination of several 3D datasets~\cite{chang2015shapenet, deitke2023objaverse, khanna2024habitat}.
Unlike reconstruction, retrieval-based modeling efficiently produces a clean and compact scene representation for downstream use without post-processing, but may introduce mismatches in geometry and appearance. 
In essence, Diorama focuses on efficient semantic matching of the observed objects under the open-world setting and understanding realistic object arrangements.
We use DuoDuoCLIP~\cite{lee2024duoduo} to encode text, images and 3D shapes into a joint embedding space.
Specifically, we propose a hierarchical retrieval strategy where we first retrieve from a text query to ensure that all 3D shape candidates lie in the same category group, then we sort retrieved shapes using an image query to re-rank shape candidates based on object appearance and take the best-matching one as the output.

\mypara{Zero-shot object pose estimation.}
We estimate the 9D pose of novel objects for semantically-matching but not necessarily exact geometrically-matching CAD objects.
Our approach is a zero-shot model-based object pose estimation method that demonstrates generalization across unseen categories and removes the requirement for pose-labeled real image training data. 
We use the semantic-rich representation of visual foundation model (\eg DinoV2~\cite{oquab2023dinov2}) to compute 2D correspondences for a pair of query image and a multiview rendering of a CAD object. We compute correspondence scores as cosine similarity between patch embeddings and take the top-K correspondences. A coarse pose hypothesis is produced by taking the most similar multiview rendering with the maximal averaged correspondence-wise similarity.
We then lift from 2D to 3D correspondences using depth. We apply the Umeyama algorithm~\cite{umeyama1991least} with RANSAC to solve for a rigid body transformation. 
We empirically find that the translation $t$ and the uniform scaling $s$ can be significantly influenced by undesired depth estimation and large object occlusion in image that result in visually implausible scene (object with extreme size or position deviation). We therefore adopt a small scale prediction network $\mathcal{F}_\text{scale}$ from GigaPose~\cite{nguyen2024gigaPose} for more robust scale estimation.
The approach is illustrated in \cref{fig:pose-estimation}.
See the supplement for more details.

\begin{figure}[t]
\centering
\includegraphics[width=0.9\linewidth]{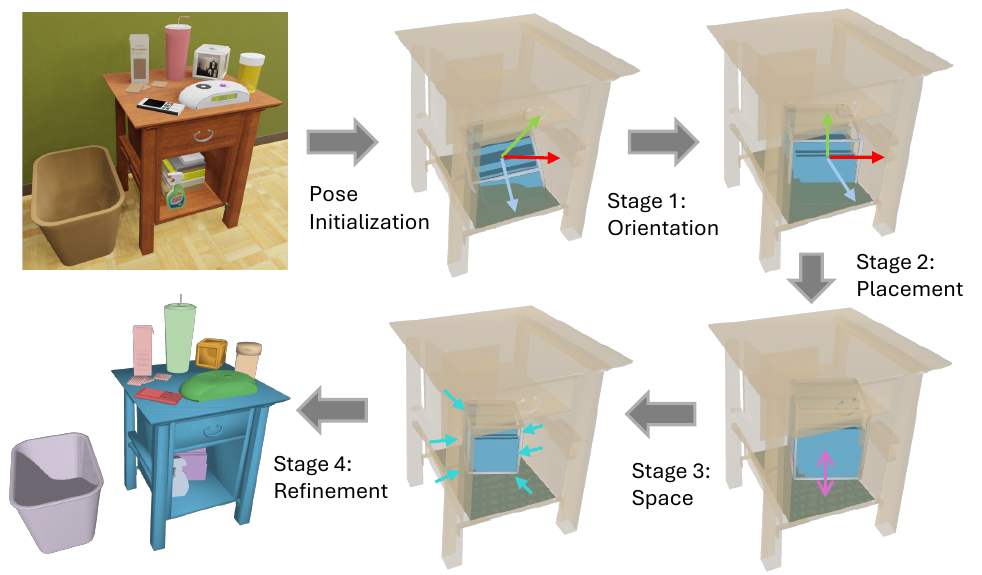}
\vspace{-6pt}
\caption{
Our semantic-aware scene layout optimization.
In this example a stack of books is placed on a side table.
} 
\vspace{-6pt}
\label{fig:layout-optimization}
\end{figure}

\mypara{Stage-wise scene optimization.}
Errors in the object pose estimation may lead to violations in overall scene plausibility.
Examples include inter-penetrations, flying objects, or gaps to the supporting objects.
To address this, we use a differentiable semantic-aware scene layout optimization procedure to refine the coarse object poses.
We separate the entire procedure into different stages to alleviate the difficulty of adjusting the 9D poses of all objects simultaneously.
For each object, we first define its oriented bounding box (OBB) and then identify potential contact and support surfaces along with contact and support directions to optimize spatial support relations with other objects.
We describe the optimization stages below and illustrate the stage-wise optimization in~\cref{fig:layout-optimization}. See the supplement for details.

\textit{Stage 1: Orientation.} 
For a pair of objects $o_i$ and $o_j$, and their associated contact surface normal $\mathbf{n}_i^c$ and support surface normal $\mathbf{n}_j^s$, we define the term $e_{\text{align}}$ as the $L_2$ norm of difference between the two normals
$
    e_{\text{align}} = \sum_{(i,j) \in G} ||\mathbf{n}_i^c - \mathbf{n}_j^s||_2
$.
During pose refinement, we maintain the front direction of each object projected on the horizontal plane by penalizing large disturbances.
Specifically, we define the term $e_{\text{sem}}$ as the $L_2$ norm of residuals added to the projected vector,
$
    e_{\text{sem}} = \sum_{i \in V} ||\mathbf{v}_i^* - \mathbf{v}_i||_2
$
where $\mathbf{v}^*$ indicates the initial front vectors before optimization.

\textit{Stage 2: Placement.}
To ensure the supported object $o_i$ is placed on the supporting object $o_j$ for physical plausibility, we define the term $e_{\text{place}}$ as the $L_1$ norm of the distance vector from the center point of the contact surface to the support surface
$
    e_{\text{place}} = \sum_{(i,j) \in G} ||D_{\text{contact center} \rightarrow \text{support surface}}||_1
$
We also define a spatial relation term $e_{\text{rel}}$ that deliberately maintains the relative $L_2$ distance between objects based on proximity relationships present in the scene,
$
    e_{\text{rel}} = \sum_{(i,j) \in G} ||(\mathbf{c}_i^* - \mathbf{c}_j^*) - (\mathbf{c}_i - \mathbf{c}_j)||_2
$.

\textit{Stage 3: Space.} Each object should occupy its own 3D space and not penetrate other objects.
In particular, we handle the usually ignored but common case where objects are placed inside another object (\eg books on a bookshelf) by defining a supporting volume for each object and an optimization term $e_{\text{vol}}$ that penalizes the summed protrusions of each object from its containing volume bounds.

\textit{Stage 4: Refinement.} Lastly, we encourage all objects to stick to their supporting surface by running the placement stage again.
We also measure the amount of collision among objects and architecture elements following~\citet{zhang2021deeppanocontext}, using the Separating Axis Theorem (SAT) to calculate the collision between convex polygons (cuboids).
We penalize the penetration depths between any two objects on three orthogonal axes, 
$
    e_{\text{col}} = \sum_{\{i,\;j\} \in V} \sum_{axis} d_
    \text{penetration}
$.

\section{Experiments}

\begin{figure}%
\centering
\includegraphics[width=\linewidth,height=5cm]{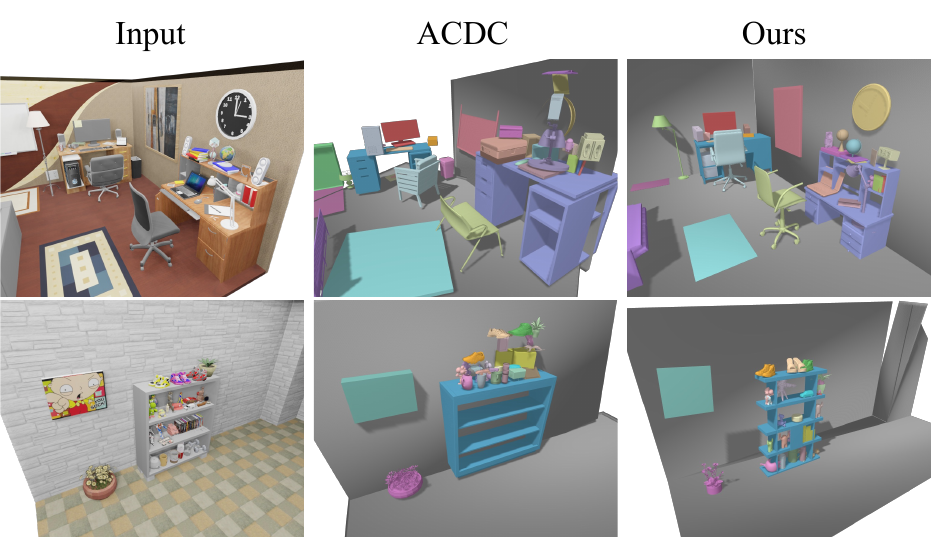}
\caption{
Comparison examples on SSDB. We capture architecture details, object geometry matching and spatial relations better.
}
\vspace{-6pt}
\label{fig:main-results}
\end{figure}

\begin{figure*}[t]
\centering
\includegraphics[width=\linewidth]{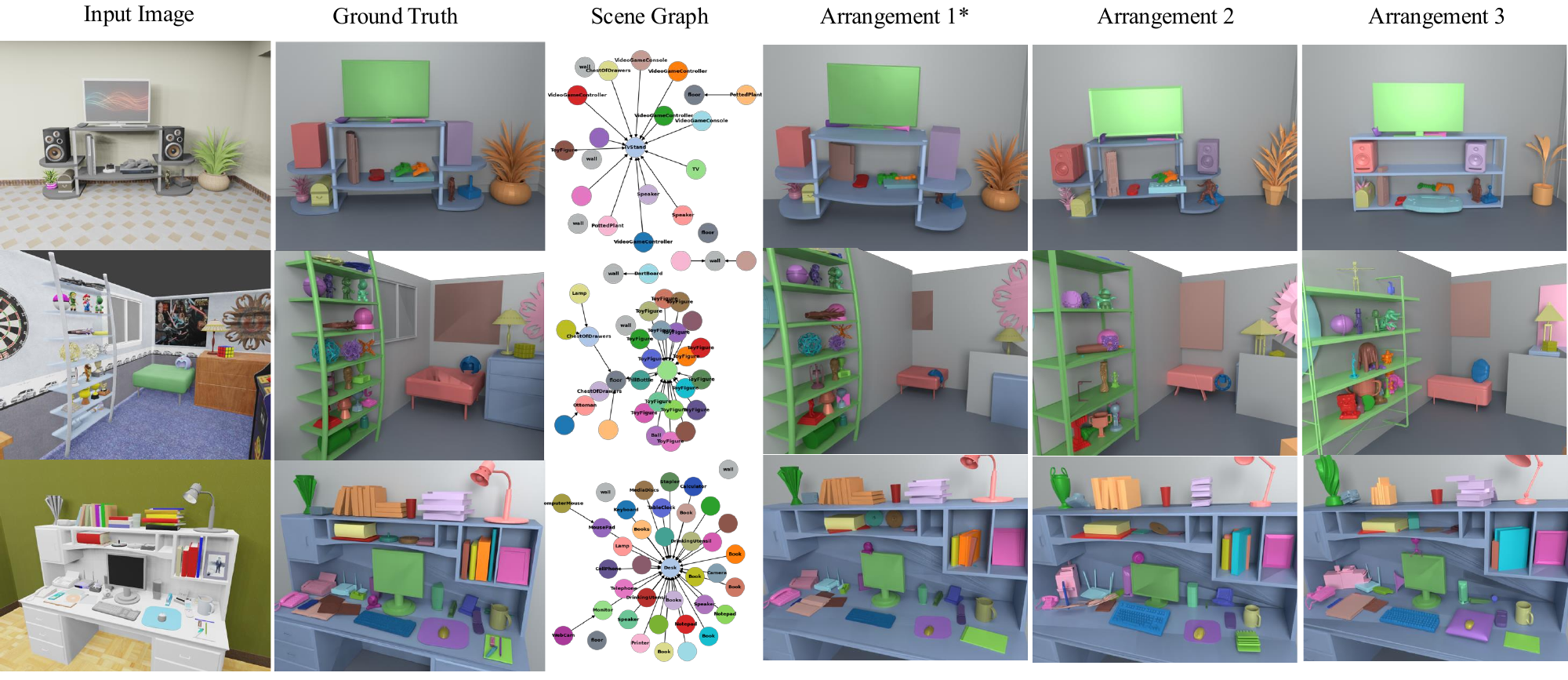}
\vspace{-16pt}
\caption{
Scene modeling examples of Diorama on SSDB images. 
The third column visualizes the support hierarchy of the generated scene graphs.
We predict object placement and arrangement in 3D.
The ``Arrangement 1$^{*}$'' column uses ground-truth 3D shapes.
``Arrangement 2/3'' presents similar semantic arrangements using different retrieved 3D shapes.
}
\vspace{-6pt}
\label{fig:wss-results}
\end{figure*}

\begin{table}[t]
\centering
\resizebox{\linewidth}{!}
{
\begin{tabular}{@{}l cccc c c cc @{}}
\toprule
\multirow{2}{*}{Method} & \multicolumn{4}{c}{Scene-aware Alignment $\uparrow$} & \multirow{2}{*}{CD $\downarrow$} & \multirow{2}{*}{User $\uparrow$} & \multicolumn{2}{c}{System Cost $\downarrow$}\\
\cmidrule(lr){2-5}  \cmidrule(lr){8-9} 
 & rAcc & tAcc & sAcc & Acc &  &  & API (\$) & Time (min) \\
\midrule
ACDC~\cite{dai2024acdc} & 0.20 & 0.56 & 0.36 & 0.04 & 14.0 & 14.9 & 1.44 & 23.2  \\
Ours & \textbf{0.23} & \textbf{0.68} & \textbf{0.49} & \textbf{0.08} & \textbf{9.5} & \textbf{85.1} & \textbf{0.12} & \textbf{3.7} \\
\bottomrule
\end{tabular}
}
\caption{System comparison on SSDB images. We exclude GT 3D shapes from retrieval for fair comparison. 
``User'' indicates the user preference over generated scenes by two methods.
}
\vspace{-6pt}
\label{tab:main_results}
\end{table}

\subsection{Experimental setup}

\mypara{Datasets.}
We demonstrate on the Stanford Scene Database (SSDB)~\cite{fisher2012example}, which contains around 130 scenes.
It provides a rich support hierarchy annotation and defines the room architecture, allowing us to evaluate our output comprehensively.
The SSDB scenes represent highly cluttered arrangements such as office desk setups and chemistry laboratories annotated using 1,722 unique CAD models ranging from large furniture (\eg chair, table, bookshelf) to small items (\eg handbag, clothing hanger, doll).
We use Blender to render 344 scene images, with 3 camera views per scene on average.
To enlarge the 3D shape repository, we include shapes from HSSD~\cite{khanna2024habitat} and the LVIS set of Objaverse~\cite{deitke2023objaverse} (Obja-LVIS) to form an out-of-distribution (OOD) collection of 60K 3D shapes for retrieval.

\mypara{Metrics.} We evaluate the different aspects of Diorama with a suite of metrics.
For architecture reconstruction, we report 2D IoU, Pixel Error (PE), Edge Error (EE) and Root Mean Squared Error (RMSE) following~\cite{stekovic2020general}. 
We also introduce bounding-box-centered chamfer distance (CDb) to measure the correctness of 3D structure of the predicted architecture.
To evaluate object arrangement, we introduce a scene-aware alignment accuracy (Acc) to eliminate the effect of inaccurate depth prediction and penalize cases where scene plausibility is severely reduced due to a few objects with unreasonable poses.
An object is considered correctly aligned under scene awareness if its relative translation deviation $\leq 20^{\circ}$ (tAcc), relative rotation error $\leq 20^{\circ}$ (rAcc) and relative scale error $\leq 0.2$ (sAcc) to \emph{every other object in the scene}.
To measure the overall scene quality, we decompose the evaluation into two aspects: average mesh collision and scene structure hierarchy where we calculate the support relation prediction accuracy (where a prediction is correct if both support object and surface assignment are correct) and supportness accuracy where whether objects are properly positioned. 
To measure 3D shape retrieval similarity against the ground-truth shape, we compute the chamfer distance (CD) as in prior work~\cite{wu2024generalizing, gao2023diffcad}, multiplied by $10^3$ for readability.

\subsection{Zero-shot Results on SSDB}

\begin{figure}[t]
\centering
\includegraphics[width=\linewidth]{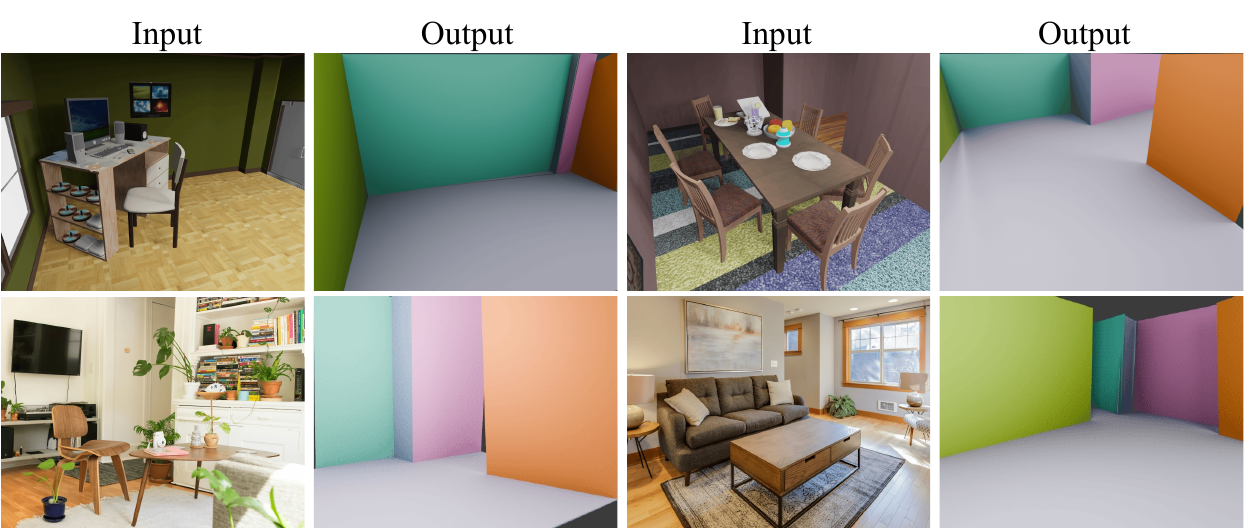}
\caption{
Examples of planar architecture reconstruction on  SSDB and real-world images.
} 
\vspace{-6pt}
\label{fig:arch-results}
\end{figure}

\begin{table}%
\centering
\resizebox{\linewidth}{!}
{
\begin{tabular}{@{} ll ccccccc @{}}
\toprule
Method & Depth & \# Success & IoU $\uparrow$ & PE $\downarrow$ & EE $\downarrow$ & RMSE $\downarrow$ & CDb $\downarrow$ & Time (s) $\downarrow$ \\
\midrule
RaC~\cite{stekovic2020general} & DAv2 & 236 & 40.3 (58.8) & 39.8 (12.2) & 23.2 & \textbf{0.67} & 0.645 & 43\\
RaC & M3Dv2 &  258 & 45.2 (60.3) & 33.7 (11.6) & 21.0 & 1.23 & 0.908 & 49 \\
ACDC~\cite{dai2024acdc} & DAv2 & 344 & 46.8 & 17.0 & 31.0 & 1.54 & 0.563 & 32 \\
PlainRecon & DAv2 &  344 & 47.8 & 13.3 & 27.5 & 1.26 & 0.503 & \textbf{23}\\
PlainRecon & M3Dv2 &  \textbf{344} & \textbf{58.6} & \textbf{9.6} & \textbf{18.9} & 1.37 & \textbf{0.447} & 29\\
\bottomrule
\end{tabular}
}
\vspace{-6pt}
\caption{
Planar architecture reconstruction results on SSDB. 
Numbers in parentheses indicate results of RaC when only considering successful predictions.
}
\vspace{-6pt}
\label{tab:arch_recon}
\end{table}

\begin{table}
\centering
\resizebox{\linewidth}{!}
{
\begin{tabular}{@{}ll rr rr rr rr rr rr @{}}
\toprule
\multirow{2}{*}{Models} & \multicolumn{2}{c}{Household} & \multicolumn{2}{c}{Furniture} & \multicolumn{2}{c}{Occluded} & \multicolumn{2}{c}{Complete} & \multicolumn{2}{c}{Supported} & \multicolumn{2}{c}{Supporting} \\
\cmidrule(lr){2-3} \cmidrule(lr){4-5} \cmidrule(lr){6-7} \cmidrule(lr){8-9} \cmidrule(lr){10-11} \cmidrule(lr){12-13}
& SS & OOD & SS & OOD & SS & OOD & SS & OOD & SS & OOD & SS & OOD \\
\midrule
CLIP-H & 8.5 & 11.2 & 5.1 & 8.8 &  8.5 & 11.3 &  7.3 & 10.2  &  8.2 & 11.1  &  6.0 & 8.9 \\
OS-H & 8.6 & 25.3 & 4.6 & 8.9 &  8.1 &  22.9 &  7.5 &  21.3 &  8.4 &  22.5 &  4.6 &  19.5  \\
DD-T &  9.4 &  12.4 &  5.9 &  10.9 &  9.1 &  12.6 &  8.3 &  11.6 &  9.2 &  12.3 &  6.5 &  10.8  \\
DD-V &  6.4 &  12.0 &  8.9 &  12.1 &  10.3 &  14.3 & \textbf{ 4.1} &  10.0 &  6.9 &  12.6 & 7.0 &   9.1 \\
DD-H (Ours) & \textbf{5.5} & \textbf{9.9} & \textbf{3.2} & \textbf{7.6} & \textbf{5.7} & \textbf{9.5} &  4.4 & \textbf{9.4} & \textbf{5.4} & \textbf{10.2} & \textbf{3.0} &  \textbf{6.1} \\
\bottomrule
\end{tabular}
}
\caption{Retrieval similarity (CD) results on SSDB images.  
We compare different models - CLIP, OpenShape (OS), DuoduoCLIP (DD) - retrieving from SSDB (SS) and out-of-distribution (OOD) objects.
``T'' and ``V'' mean using text or visual modality for query only respectively. ``H'' stands for using both modalities.
}
\vspace{-6pt}
\label{tab:wss_retrieval}
\end{table}

\begin{table}%
\centering
\resizebox{\linewidth}{!}
{
\begin{tabular}{@{}l cccccc @{}}
\toprule
\multirow{2}{*}{Method}  & \multicolumn{4}{c}{Scene-aware Alignment $\uparrow$} & \multirow{2}{*}{Collision $\downarrow$} & \multirow{2}{*}{Relation $\uparrow$} \\
\cmidrule(lr){2-5} 
& rAcc & tAcc & sAcc & Acc &  &   \\
\midrule
BM baseline & 0.34 & 0.93 & 0.52 & 0.19 & 11.43  & 0.58  \\
ZSP~\cite{goodwin2022zero} & 0.36 & 0.92 & 0.59 & 0.25 & 8.76 & 0.60  \\
ZSP w/ DinoV2 & 0.37 & 0.85 & 0.56 & 0.26 & 9.59 & 0.59  \\
ZSP w/ ft DinoV2 & 0.38 & 0.93 & 0.58 & 0.25 & 10.58 & 0.44  \\
GigaPose~\cite{nguyen2024gigaPose} & 0.36 & 0.95 & \textbf{0.71} & 0.27 & 7.91  & 0.61  \\
Ours & \textbf{0.47} & \textbf{0.95} & 0.70 & \textbf{0.37} & \textbf{6.42} & \textbf{0.62} \\
\bottomrule
\end{tabular}
}
\vspace{-6pt}
\caption{
Comparison of zero-shot pose estimation methods for the 9-DoF CAD alignment task on SSDB images.
GT objects and depth are used to avoid 2D perception error for system analysis.
}
\label{tab:wss_pose_gtd}
\end{table}

\begin{table}[t]
\centering
\resizebox{\linewidth}{!}
{
\begin{tabular}{@{}l c ccc @{}}
\toprule
\multirow{2}{*}{Ablation} & \multirow{2}{*}{Collision $\downarrow$} & \multicolumn{3}{c}{Scene Structure $\uparrow$}  \\
\cmidrule(lr){3-5} 
 &  & orientation & placement & overall \\
\midrule
w/o optimization & 6.42 & 0.19 & 0.01 & 0.00 \\
\midrule
+ \textbf{O}rientation & 5.17 & 0.98 & 0.01 & 0.01  \\
+ O, \textbf{P}lacement & 8.08 & 0.98 & 0.93 & 0.91 \\
+ O, P, \textbf{S}pace & 6.08 & 0.98 & 0.54 & 0.54  \\
+ O, P, S, \textbf{R}efinement & \textbf{3.78} & \textbf{0.98} & \textbf{0.95} & \textbf{0.93} \\
\midrule
w/ GT scene graph & 3.12 & 0.98 & 0.94 & 0.92 \\
w/ GT architecture & 3.88 & 0.96 & 0.93 & 0.90 \\
\bottomrule
\end{tabular}
}
\caption{Ablation of the terms in our semantic-aware scene optimization.
Scene graphs are generated with GPT-4o.
}
\vspace{-6pt}
\label{tab:wss_lo}
\end{table}

\begin{figure}[t]
\centering
\includegraphics[width=\linewidth]{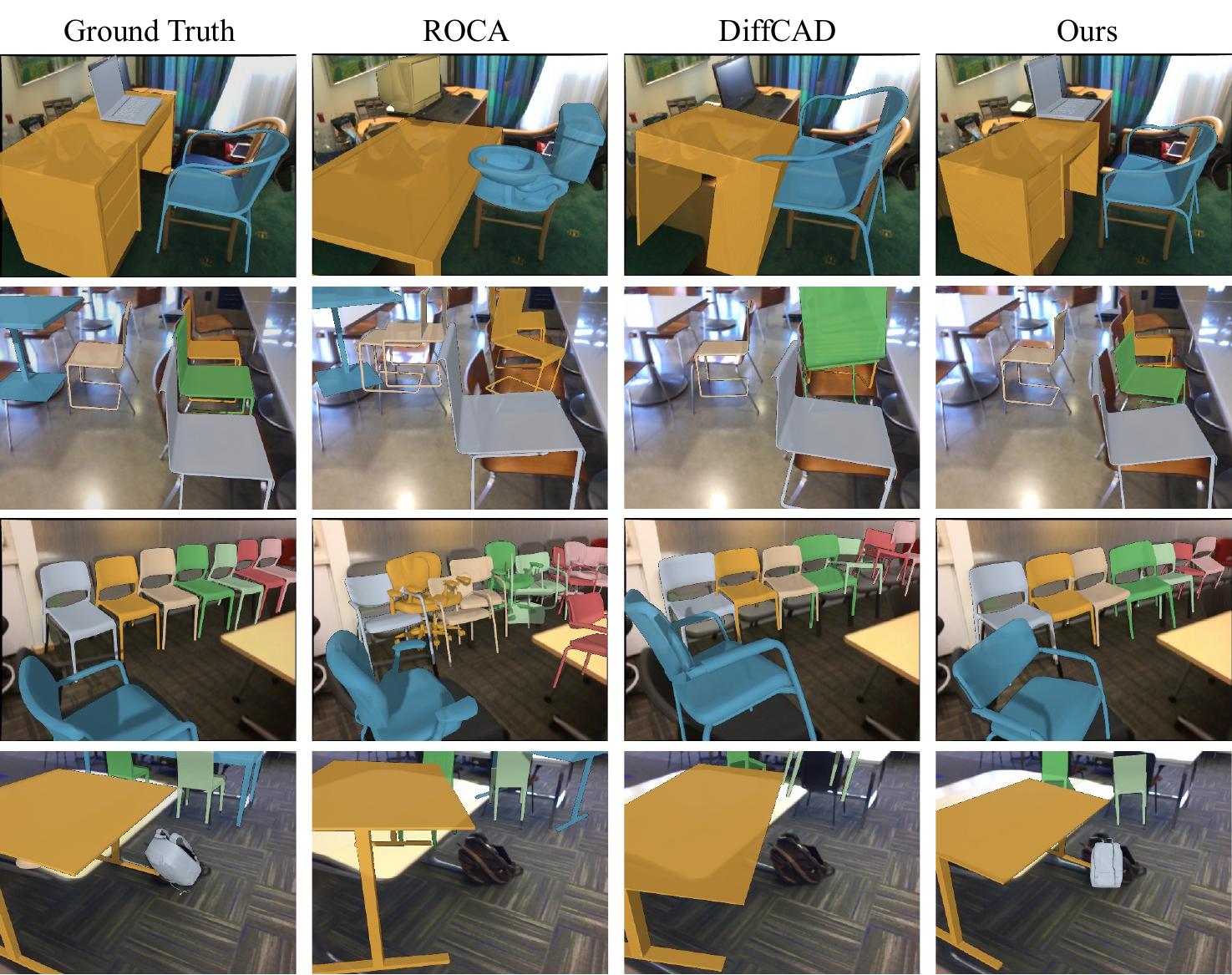}
\caption{
Examples on ScanNet images. 
We use the same mask, depth and 3D shape inputs for DiffCAD and ours for fair comparison as they are both modular methods.
We handle more categories in 1st and 4th examples (laptop and backpack).
} 
\vspace{-6pt}
\label{fig:scan2cad}
\end{figure}

\begin{figure*}%
\centering
\setkeys{Gin}{width=\linewidth}
\setlength{\tabcolsep}{2pt}
\begin{tabularx}{\linewidth}{@{} XXXXXX @{}}
\multicolumn{1}{c}{\scriptsize Input} & \multicolumn{1}{c}{\scriptsize Output} & \multicolumn{1}{c}{\scriptsize Input} & \multicolumn{1}{c}{\scriptsize Output} & \multicolumn{1}{c}{\scriptsize Input} & \multicolumn{1}{c}{\scriptsize Output}\\
\imgclip{0}{fig/images/wild_figure_supp/scene_15_gt} & \imgclip{0}{fig/images/wild_figure_supp/scene_15_diorama}
 & \imgclip{0}{fig/images/wild_figure/scene_09_gt} & \imgclip{0}{fig/images/wild_figure/scene_09_diorama} & \imgclip{0}{fig/images/wild_figure_supp/scene_24_gt} & \imgclip{0}{fig/images/wild_figure_supp/scene_24_diorama} \\
\imgclip{0}{fig/images/wild_figure/scene_44_gt} & \imgclip{0}{fig/images/wild_figure/scene_44_diorama} & \imgclip{0}{fig/images/wild_figure/scene_52_gt} & \imgclip{0}{fig/images/wild_figure/scene_52_diorama} & \imgclip{0}{fig/images/wild_figure_supp/scene_56_gt} & \imgclip{0}{fig/images/wild_figure_supp/scene_56_diorama}  \\
\end{tabularx}
\vspace{-6pt}
\caption{
Examples on real-world internet images.
The output scenes capture large and small objects, and represent complex relations such as objects on walls (e.g., wall clock in top left) and objects on other objects (vase with flower on table, in the top right).
}
\vspace{-6pt}
\label{fig:real}
\end{figure*}

\begin{figure}%
\centering
\setkeys{Gin}{width=\linewidth}
\setlength{\tabcolsep}{2pt}
\begin{tabularx}{\linewidth}{@{} XXX @{}}
\multicolumn{1}{c}{\scriptsize Text prompt} & \multicolumn{1}{c}{\scriptsize Generated image} & \multicolumn{1}{c}{\scriptsize Output} \\
\imgclip{0}{fig/images/flux_figure_supp/scene_08_prompt} & \imgclip{0}{fig/images/flux_figure_supp/scene_08_gt} & \imgclip{0}{fig/images/flux_figure_supp/scene_08_diorama}  \\
\imgclip{0}{fig/images/flux_figure/scene_105_prompt}& \imgclip{0}{fig/images/flux_figure/scene_105_gt} & \imgclip{0}{fig/images/flux_figure/scene_105_diorama}  \\
\end{tabularx}
\vspace{-6pt}
\caption{
Examples of applying Diorama in a text-to-scene setting.
The output scenes conform to the text prompt.
}
\vspace{-6pt}
\label{fig:text2scene}
\end{figure}

We present evaluations of overall system performance, architecture reconstruction, shape retrieval, object pose estimation and scene optimization.
See the supplement for more ablations.
\cref{fig:wss-results} shows multiple semantic-similar arrangements produced by Diorama on scenes of complex layout hierarchy.

\mypara{System comparison.} 
We compare with ACDC~\cite{dai2024acdc}, another modular system of building synthetic scenes from images for robot policy learning. We evaluate the overall scene modeling performance over metrics of object alignment, similarity and system cost.
We exclude ground-truth SSDB 3D shapes from retrieval for fair comparison.
\cref{tab:main_results} shows that Diorama outperforms ACDC noticeably.
Both methods present relatively low performance on alignment accuracy due to object retrieval mismatch and undesired depth prediction.
While both approaches leverage LLM, ours is $\times12$ cheaper and $\times6$ faster than ACDC. 
We also conduct a user study in terms of object matching and scene quality on sampled reconstructed scenes. We achieve 85.1\% user preference on outputs generated by our method.
We show qualitative examples in~\cref{fig:main-results} and note that Diorama reconstructs more complete and fine-grained room structure, and captures better object arrangements. Since ACDC heavily relies on a physics engine to solve the collision and floating problem, it cannot explicitly place an object on certain supporting surface. See more examples in the supplement.

\mypara{Architecture plane reconstruction.}
We compare our proposed PlainRecon with RaC and ACDC in \cref{tab:arch_recon}, with predicted depth from DepthAnythingV2 (DAv2) and Metric3DV2 (M3Dv2).
We find that RaC fails in roughly 30\% of the cases due to infeasible solutions produced by the constrained discrete optimization procedure, while PlainRecon and ACDC are able to handle all inputs. 
PlainRecon remarkably outperforms RaC and ACDC on most of the metrics, while RMSE being an outlier where PlainRecon underperforms significantly.
The superior performance of PlainRecon on CDb means that it captures the 3D room structure of the scene better than others.
Our results also show that M3Dv2 depth is overall superior to DAv2 for architecture reconstruction. 
We visualize both reconstruction results on SSDB and internet images in~\cref{fig:arch-results}.

\mypara{3D shape retrieval results.}
\cref{tab:wss_retrieval} evaluates our multimodal 3D shape retrieval approach against CLIP~\cite{radford2021learning} and OpenShape~\cite{liu2024openshape}.
We use the ViT-bigG-14 checkpoint of CLIP.
We present results both for retrieving from SSDB shapes, and from a larger set of OOD shapes.
We retrieve top-5 objects and take the candidate with the minimum L1 Chamfer Distance to the ground truth.
Our approach outperforms others by a big margin reflecting that multiple modalities compensate for potential ambiguity in language or visual information loss in a photo of a complicated scene.
The inferior performance of OpenShape implies that the multiview representation of 3D shapes we use is a better fit for the open-world problem setting.

\mypara{9D CAD alignment to image object.}
We compare our zero-shot pose estimation module with representative methods, including 
ZSP~\cite{goodwin2022zero} and GigaPose~\cite{nguyen2024gigaPose}, on the 9-DoF CAD alignment task on SSDB images. Since GigaPose originally cannot predict the size of CAD model, we augment it by integrating depth information. 
We include a baseline (BM) which only performs coarse pose selection using the best-matching multiview image with the highest average correspondence similarity.
For ZSP, we experiment with variants in terms of different feature extractors, including the default Dino~\cite{caron2021emerging}, DinoV2 and a fine-tuned DinoV2 backbone from GigaPose. 
\cref{tab:wss_pose_gtd} shows that our method performs the best as measured by metrics for overall alignment, collision and support relation prediction, while achieving comparable performance on sAcc against GigaPose. 
The failure on scale estimation is due to the RANSAC-based pose solver that is substantially affected by noisy point correspondences. See additional results on estimated depth in the supplement.

\mypara{Benefits of semantic-aware scene optimization.}
We conduct an ablation of the terms in the semantic-aware scene optimization.
We set more strict thresholds for alignment accuracy where translation error $\leq$ 0.1cm and geodesic rotation error $\leq 1^{\circ}$.
\cref{tab:wss_lo} shows that the full scene optimization (+ O,P,S,R) reduces collisions avoidance and remarkably improves scene hierarchical structure. The ablated results of gradually adding more optimization terms show that the orientation (O), placement (P), space (S) and refinement (R) terms all help improve object alignment.
This table also reports results with a pseudo ground-truth scene graph and architecture to show that improvements to these modules can lead to better overall scene modeling.

\subsection{Zero-shot Results on ScanNet}

We validate our zero-shot open-world system on real-world images from ScanNet25k~\cite{dai2017scannet} and ShapeNet~\cite{chang2015shapenet} CAD model annotations provided by Scan2CAD~\cite{avetisyan2019scan2cad}.
We deliberately evaluate on cluttered environments capturing occlusion and multiple objects.
We select 600 images from the standard evaluation set where each image contains at least 3 different object categories or 5 object instances.
These images contain around 3,000 instances spanning 24 categories, compared to the evaluation set in prior work~\cite{gumeli2022roca} with 2,500 instances from 2,100 images covering 6 categories.
We compare to ROCA~\cite{gumeli2022roca} and DiffCAD~\cite{gao2023diffcad} where ROCA is a fully-supervised deterministic end-to-end model trained solely on Scan2CAD and DiffCAD is a weakly-supervised probabilistic model trained on generated synthetic data.
Both prior works cannot scale up and generalize to unobserved real-world images due to expensive data acquisition, infeasibility of training individual models for every category or inefficient inference.
In contrast, Diorama provides a training-free solution showing comparable performance against DiffCAD and more flexible capability against ROCA.
We show a qualitative comparison of CAD alignment in~\cref{fig:scan2cad}.
Note that we can handle out-of-distribution objects (laptop and backpack) in the first and fourth row, which ROCA and DiffCAD cannot process due to limited data.
See the supplement for more results.

\subsection{Zero-shot Results on in-the-wild Images}

\mypara{Real-world internet images.}
We select a set of real-world internet images representing real-world object arrangement from \href{https://pixabay.com/}{pixabay},  \href{https://www.pexels.com/}{Pexels} and \href{https://unsplash.com}{Unsplash}.
See \cref{fig:real} and more examples in the supplement.

\mypara{Text-conditioned generated images.}
We further expand the capability of Diorama to the text-to-scene problem. We first generate high-resolution realistic image given a text prompt using Flux-1 and apply our system to the image. See \cref{fig:text2scene} and more examples in the supplement.

\section{Conclusion}

\mypara{Limitations.} Open-world 3D perception through CAD model retrieval and alignment creates a compact 3D scene representation.
However, applications requiring precise reconstructions may be hindered by the lack of exact geometry and texture matches.
To address this, future work can investigate deforming and texturing the retrieved CAD models to improve alignment with observations while maintaining compactness, part-level structure and functionality.
More advanced methods can be explored to learn affordance between objects for proper support. 
As RANSAC-based pose estimation can be inefficient and sensitive to noisy correspondences and large occlusions, stronger open-vocabulary pose estimation methods can be more preferable for real-time scene modeling applications.

We introduced Diorama, the first zero-shot open-world system that can holistically model a 3D scene from an RGB image without requiring end-to-end training or human annotations.
Our system is composed of robust, generalizable solutions to each subtask of our problem: architecture reconstruction, 3D shape retrieval, 9D object pose estimation, and scene layout optimization. We evaluate our system on synthetic images and real-world data with quantitative analysis and visual comparisons, and demonstrate generalization to real-world images and text-to-scene applications.

\mypara{Acknowledgements.}
This work was funded in part by a CIFAR AI Chair, a Canada Research Chair, NSERC Discovery Grant, and NSF award \#2016532. Daniel Ritchie is an advisor to Geopipe and owns equity in the company. Geopipe is a start-up that is developing 3D technology to build immersive virtual copies of the real world with applications in various fields, including games and architecture. 
We thank Xiaohao Sun, Jiayi Liu, Xingguang Yan, Chirag Vashist, Wuyang Chen, Dongchen Yang and Mahdi Miangoleh for feedback on the paper draft.

{
\small
\bibliographystyle{ieeenat_fullname}
\bibliography{main}

\begin{thebibliography}{99}
\providecommand{\natexlab}[1]{#1}
\providecommand{\url}[1]{\texttt{#1}}
\expandafter\ifx\csname urlstyle\endcsname\relax
  \providecommand{\doi}[1]{doi: #1}\else
  \providecommand{\doi}{doi: \begingroup \urlstyle{rm}\Url}\fi

\bibitem[Abdelreheem et~al.(2023)Abdelreheem, Skorokhodov, Ovsjanikov, and Wonka]{abdelreheem2023satr}
Ahmed Abdelreheem, Ivan Skorokhodov, Maks Ovsjanikov, and Peter Wonka.
\newblock {SATR}: Zero-shot semantic segmentation of {3D} shapes.
\newblock In \emph{Proceedings of the IEEE/CVF International Conference on Computer Vision}, pages 15166--15179, 2023.

\bibitem[Agarwal et~al.(2024)Agarwal, Singh, Sen, Lozano-P{\'e}rez, and Kaelbling]{agarwal2024scenecomplete}
Aditya Agarwal, Gaurav Singh, Bipasha Sen, Tom{\'a}s Lozano-P{\'e}rez, and Leslie~Pack Kaelbling.
\newblock {SceneComplete}: Open-world {3D} scene completion in complex real world environments for robot manipulation.
\newblock \emph{arXiv preprint arXiv:2410.23643}, 2024.

\bibitem[Aguina-Kang et~al.(2024)Aguina-Kang, Gumin, Han, Morris, Yoo, Ganeshan, Jones, Wei, Fu, and Ritchie]{aguina2024open}
Rio Aguina-Kang, Maxim Gumin, Do~Heon Han, Stewart Morris, Seung~Jean Yoo, Aditya Ganeshan, R~Kenny Jones, Qiuhong~Anna Wei, Kailiang Fu, and Daniel Ritchie.
\newblock Open-universe indoor scene generation using {LLM} program synthesis and uncurated object databases.
\newblock \emph{arXiv preprint arXiv:2403.09675}, 2024.

\bibitem[Avetisyan et~al.(2019)Avetisyan, Dahnert, Dai, Savva, Chang, and Nie{\ss}ner]{avetisyan2019scan2cad}
Armen Avetisyan, Manuel Dahnert, Angela Dai, Manolis Savva, Angel~X Chang, and Matthias Nie{\ss}ner.
\newblock {Scan2CAD}: Learning {CAD} model alignment in {RGB-D} scans.
\newblock In \emph{Proceedings of the IEEE/CVF Conference on computer vision and pattern recognition}, pages 2614--2623, 2019.

\bibitem[Caron et~al.(2021)Caron, Touvron, Misra, J\'egou, Mairal, Bojanowski, and Joulin]{caron2021emerging}
Mathilde Caron, Hugo Touvron, Ishan Misra, Herv\'e J\'egou, Julien Mairal, Piotr Bojanowski, and Armand Joulin.
\newblock Emerging properties in self-supervised vision transformers.
\newblock In \emph{Proceedings of the International Conference on Computer Vision (ICCV)}, 2021.

\bibitem[Chang et~al.(2015)Chang, Funkhouser, Guibas, Hanrahan, Huang, Li, Savarese, Savva, Song, Su, et~al.]{chang2015shapenet}
Angel~X Chang, Thomas Funkhouser, Leonidas Guibas, Pat Hanrahan, Qixing Huang, Zimo Li, Silvio Savarese, Manolis Savva, Shuran Song, Hao Su, et~al.
\newblock {ShapeNet}: An information-rich {3D} model repository.
\newblock \emph{arXiv preprint arXiv:1512.03012}, 2015.

\bibitem[Chen et~al.(2023)Chen, Liu, Kong, Zhu, Ma, Li, Hou, Qiao, and Wang]{chen2023clip2scene}
Runnan Chen, Youquan Liu, Lingdong Kong, Xinge Zhu, Yuexin Ma, Yikang Li, Yuenan Hou, Yu Qiao, and Wenping Wang.
\newblock {CLIP2Scene}: Towards label-efficient {3D} scene understanding by {CLIP}.
\newblock In \emph{Proceedings of the IEEE/CVF Conference on Computer Vision and Pattern Recognition}, pages 7020--7030, 2023.

\bibitem[Chen et~al.(2024{\natexlab{a}})Chen, Ni, Jiang, Zhang, Zhu, and Huang]{chen2024single}
Yixin Chen, Junfeng Ni, Nan Jiang, Yaowei Zhang, Yixin Zhu, and Siyuan Huang.
\newblock Single-view {3D} scene reconstruction with high-fidelity shape and texture.
\newblock In \emph{2024 International Conference on 3D Vision (3DV)}, pages 1456--1467. IEEE, 2024{\natexlab{a}}.

\bibitem[Chen et~al.(2024{\natexlab{b}})Chen, Walsman, Memmel, Mo, Fang, Vemuri, Wu, Fox, and Gupta]{chen2024urdformer}
Zoey Chen, Aaron Walsman, Marius Memmel, Kaichun Mo, Alex Fang, Karthikeya Vemuri, Alan Wu, Dieter Fox, and Abhishek Gupta.
\newblock Urdformer: A pipeline for constructing articulated simulation environments from real-world images.
\newblock \emph{arXiv preprint arXiv:2405.11656}, 2024{\natexlab{b}}.

\bibitem[Dai et~al.(2017)Dai, Chang, Savva, Halber, Funkhouser, and Nie{\ss}ner]{dai2017scannet}
Angela Dai, Angel~X Chang, Manolis Savva, Maciej Halber, Thomas Funkhouser, and Matthias Nie{\ss}ner.
\newblock {ScanNet}: Richly-annotated {3D} reconstructions of indoor scenes.
\newblock In \emph{Proceedings of the IEEE conference on computer vision and pattern recognition}, pages 5828--5839, 2017.

\bibitem[Dai et~al.(2024)Dai, Wong, Jiang, Wang, Gokmen, Zhang, Wu, and Fei-Fei]{dai2024acdc}
Tianyuan Dai, Josiah Wong, Yunfan Jiang, Chen Wang, Cem Gokmen, Ruohan Zhang, Jiajun Wu, and Li Fei-Fei.
\newblock {ACDC}: Automated creation of digital cousins for robust policy learning.
\newblock \emph{arXiv preprint arXiv:2410.07408}, 2024.

\bibitem[Deitke et~al.(2023)Deitke, Schwenk, Salvador, Weihs, Michel, VanderBilt, Schmidt, Ehsani, Kembhavi, and Farhadi]{deitke2023objaverse}
Matt Deitke, Dustin Schwenk, Jordi Salvador, Luca Weihs, Oscar Michel, Eli VanderBilt, Ludwig Schmidt, Kiana Ehsani, Aniruddha Kembhavi, and Ali Farhadi.
\newblock Objaverse: A universe of annotated 3d objects.
\newblock In \emph{Proceedings of the IEEE/CVF Conference on Computer Vision and Pattern Recognition}, pages 13142--13153, 2023.

\bibitem[Delitzas et~al.(2024)Delitzas, Takmaz, Tombari, Sumner, Pollefeys, and Engelmann]{delitzas2024scenefun3d}
Alexandros Delitzas, Ayca Takmaz, Federico Tombari, Robert Sumner, Marc Pollefeys, and Francis Engelmann.
\newblock {SceneFun3D}: Fine-grained functionality and affordance understanding in {3D} scenes.
\newblock In \emph{IEEE/CVF Conference on Computer Vision and Pattern Recognition (CVPR)}, 2024.

\bibitem[Ding et~al.(2023)Ding, Yang, Xue, Zhang, Bai, and Qi]{ding2023pla}
Runyu Ding, Jihan Yang, Chuhui Xue, Wenqing Zhang, Song Bai, and Xiaojuan Qi.
\newblock {PLA}: Language-driven open-vocabulary {3D} scene understanding.
\newblock In \emph{Proceedings of the IEEE/CVF conference on computer vision and pattern recognition}, pages 7010--7019, 2023.

\bibitem[Dong et~al.(2024)Dong, Fang, Bo, Dong, and Tan]{dong2024panocontext}
Yuan Dong, Chuan Fang, Liefeng Bo, Zilong Dong, and Ping Tan.
\newblock {PanoContext-Former}: Panoramic total scene understanding with a transformer.
\newblock In \emph{Proceedings of the IEEE/CVF Conference on Computer Vision and Pattern Recognition}, pages 28087--28097, 2024.

\bibitem[Ekin et~al.(2024)Ekin, Yildirim, Caglar, Erdem, Erdem, and Dundar]{ekin2024clipaway}
Yi{\u{g}}it Ekin, Ahmet~Burak Yildirim, Erdem~Eren Caglar, Aykut Erdem, Erkut Erdem, and Aysegul Dundar.
\newblock {CLIPA}way: Harmonizing focused embeddings for removing objects via diffusion models.
\newblock In \emph{Advances in Neural Information Processing Systems}, 2024.

\bibitem[El~Banani et~al.(2024)El~Banani, Raj, Maninis, Kar, Li, Rubinstein, Sun, Guibas, Johnson, and Jampani]{el2024probing}
Mohamed El~Banani, Amit Raj, Kevis-Kokitsi Maninis, Abhishek Kar, Yuanzhen Li, Michael Rubinstein, Deqing Sun, Leonidas Guibas, Justin Johnson, and Varun Jampani.
\newblock Probing the {3D} awareness of visual foundation models.
\newblock In \emph{Proceedings of the IEEE/CVF Conference on Computer Vision and Pattern Recognition}, pages 21795--21806, 2024.

\bibitem[Ester et~al.(1996)Ester, Kriegel, Sander, and Xu]{ester1996dbscan}
Martin Ester, Hans-Peter Kriegel, J\"{o}rg Sander, and Xiaowei Xu.
\newblock A density-based algorithm for discovering clusters in large spatial databases with noise.
\newblock In \emph{Proceedings of the Second International Conference on Knowledge Discovery and Data Mining}, 1996.

\bibitem[Fischler and Bolles(1981)]{fischler1981ransac}
Martin~A. Fischler and Robert~C. Bolles.
\newblock Random sample consensus: a paradigm for model fitting with applications to image analysis and automated cartography.
\newblock \emph{Communications of the ACM}, 24\penalty0 (6):\penalty0 381--395, 1981.

\bibitem[Fisher et~al.(2012)Fisher, Ritchie, Savva, Funkhouser, and Hanrahan]{fisher2012example}
Matthew Fisher, Daniel Ritchie, Manolis Savva, Thomas Funkhouser, and Pat Hanrahan.
\newblock Example-based synthesis of 3d object arrangements.
\newblock \emph{ACM Transactions on Graphics (TOG)}, 31\penalty0 (6):\penalty0 1--11, 2012.

\bibitem[Fu et~al.(2021)Fu, Cai, Gao, Zhang, Wang, Li, Zeng, Sun, Jia, Zhao, et~al.]{fu20213d}
Huan Fu, Bowen Cai, Lin Gao, Ling-Xiao Zhang, Jiaming Wang, Cao Li, Qixun Zeng, Chengyue Sun, Rongfei Jia, Binqiang Zhao, et~al.
\newblock 3d-front: 3d furnished rooms with layouts and semantics.
\newblock In \emph{Proceedings of the IEEE/CVF International Conference on Computer Vision}, pages 10933--10942, 2021.

\bibitem[Fu et~al.(2024)Fu, Liu, Chen, Nie, and Xiong]{fu2024scene}
Rao Fu, Jingyu Liu, Xilun Chen, Yixin Nie, and Wenhan Xiong.
\newblock {Scene-LLM}: Extending language model for {3D} visual understanding and reasoning.
\newblock \emph{arXiv preprint arXiv:2403.11401}, 2024.

\bibitem[Fu et~al.(2025)Fu, Wen, Liu, and Sridhar]{fu2025anyhome}
Rao Fu, Zehao Wen, Zichen Liu, and Srinath Sridhar.
\newblock Anyhome: Open-vocabulary generation of structured and textured 3d homes.
\newblock In \emph{European Conference on Computer Vision}, pages 52--70. Springer, 2025.

\bibitem[Gao et~al.(2024{\natexlab{a}})Gao, Rozenberszki, Leutenegger, and Dai]{gao2023diffcad}
Daoyi Gao, David Rozenberszki, Stefan Leutenegger, and Angela Dai.
\newblock {DiffCAD}: Weakly-supervised probabilistic {CAD} model retrieval and alignment from an {RGB} image.
\newblock \emph{ACM Transactions on Graphics (TOG)}, 43\penalty0 (4):\penalty0 1--15, 2024{\natexlab{a}}.

\bibitem[Gao et~al.(2024{\natexlab{b}})Gao, Liu, Chen, Geiger, and Sch{\"o}lkopf]{gao2024graphdreamer}
Gege Gao, Weiyang Liu, Anpei Chen, Andreas Geiger, and Bernhard Sch{\"o}lkopf.
\newblock {GraphDreamer}: Compositional {3D} scene synthesis from scene graphs.
\newblock In \emph{Proceedings of the IEEE/CVF Conference on Computer Vision and Pattern Recognition}, pages 21295--21304, 2024{\natexlab{b}}.

\bibitem[Goodwin et~al.(2022)Goodwin, Vaze, Havoutis, and Posner]{goodwin2022zero}
Walter Goodwin, Sagar Vaze, Ioannis Havoutis, and Ingmar Posner.
\newblock Zero-shot category-level object pose estimation.
\newblock In \emph{European Conference on Computer Vision}, pages 516--532. Springer, 2022.

\bibitem[G{\"u}meli et~al.(2022)G{\"u}meli, Dai, and Nie{\ss}ner]{gumeli2022roca}
Can G{\"u}meli, Angela Dai, and Matthias Nie{\ss}ner.
\newblock {ROCA}: Robust {CAD} model retrieval and alignment from a single image.
\newblock In \emph{Proceedings of the IEEE Conference on Computer Vision and Pattern Recognition (CVPR)}, 2022.

\bibitem[Hong et~al.(2023)Hong, Zhen, Chen, Zheng, Du, Chen, and Gan]{hong20233d}
Yining Hong, Haoyu Zhen, Peihao Chen, Shuhong Zheng, Yilun Du, Zhenfang Chen, and Chuang Gan.
\newblock {3D-LLM}: Injecting the {3D} world into large language models.
\newblock \emph{Advances in Neural Information Processing Systems}, 36:\penalty0 20482--20494, 2023.

\bibitem[Hu et~al.(2024)Hu, Yin, Zhang, Cai, Long, Chen, Wang, Yu, Shen, and Shen]{hu2024metric3d}
Mu Hu, Wei Yin, Chi Zhang, Zhipeng Cai, Xiaoxiao Long, Hao Chen, Kaixuan Wang, Gang Yu, Chunhua Shen, and Shaojie Shen.
\newblock {Metric3D} v2: A versatile monocular geometric foundation model for zero-shot metric depth and surface normal estimation.
\newblock \emph{arXiv preprint arXiv:2404.15506}, 2024.

\bibitem[Huang et~al.(2023{\natexlab{a}})Huang, Krishna, Atekha, and Guibas]{huang2023aladdin}
Ian Huang, Vrishab Krishna, Omoruyi Atekha, and Leonidas Guibas.
\newblock Aladdin: Zero-shot hallucination of stylized {3D} assets from abstract scene descriptions.
\newblock \emph{arXiv preprint arXiv:2306.06212}, 2023{\natexlab{a}}.

\bibitem[Huang et~al.(2018)Huang, Qi, Zhu, Xiao, Xu, and Zhu]{huang2018holistic}
Siyuan Huang, Siyuan Qi, Yixin Zhu, Yinxue Xiao, Yuanlu Xu, and Song-Chun Zhu.
\newblock Holistic {3D} scene parsing and reconstruction from a single {RGB} image.
\newblock In \emph{Proceedings of the European conference on computer vision (ECCV)}, pages 187--203, 2018.

\bibitem[Huang et~al.(2023{\natexlab{b}})Huang, Wu, Chen, Zhao, Zhu, and Lasenby]{huang2023openins3d}
Zhening Huang, Xiaoyang Wu, Xi Chen, Hengshuang Zhao, Lei Zhu, and Joan Lasenby.
\newblock {OpenIns3D}: Snap and lookup for {3D} open-vocabulary instance segmentation.
\newblock \emph{arXiv preprint arXiv:2309.00616}, 2023{\natexlab{b}}.

\bibitem[Irshad et~al.(2022)Irshad, Kollar, Laskey, Stone, and Kira]{irshad2022centersnap}
Muhammad~Zubair Irshad, Thomas Kollar, Michael Laskey, Kevin Stone, and Zsolt Kira.
\newblock {CenterSnap}: Single-shot multi-object {3D} shape reconstruction and categorical {6D} pose and size estimation.
\newblock In \emph{2022 International Conference on Robotics and Automation (ICRA)}, pages 10632--10640. IEEE, 2022.

\bibitem[Izadinia et~al.(2017)Izadinia, Shan, and Seitz]{izadinia2017im2cad}
Hamid Izadinia, Qi Shan, and Steven~M Seitz.
\newblock Im2cad.
\newblock In \emph{Proceedings of the IEEE conference on computer vision and pattern recognition}, pages 5134--5143, 2017.

\bibitem[Jatavallabhula et~al.(2023)Jatavallabhula, Kuwajerwala, Gu, Omama, Chen, Maalouf, Li, Iyer, Saryazdi, Keetha, et~al.]{jatavallabhula2023conceptfusion}
Krishna~Murthy Jatavallabhula, Alihusein Kuwajerwala, Qiao Gu, Mohd Omama, Tao Chen, Alaa Maalouf, Shuang Li, Ganesh Iyer, Soroush Saryazdi, Nikhil Keetha, et~al.
\newblock {ConceptFusion}: Open-set multimodal {3D} mapping.
\newblock \emph{arXiv preprint arXiv:2302.07241}, 2023.

\bibitem[Jia et~al.(2025)Jia, Chen, Yu, Wang, Niu, Liu, Li, and Huang]{jia2025sceneverse}
Baoxiong Jia, Yixin Chen, Huangyue Yu, Yan Wang, Xuesong Niu, Tengyu Liu, Qing Li, and Siyuan Huang.
\newblock {SceneVerse}: Scaling {3D} vision-language learning for grounded scene understanding.
\newblock In \emph{European Conference on Computer Vision}, pages 289--310. Springer, 2025.

\bibitem[Kerr et~al.(2023)Kerr, Kim, Goldberg, Kanazawa, and Tancik]{kerr2023lerf}
Justin Kerr, Chung~Min Kim, Ken Goldberg, Angjoo Kanazawa, and Matthew Tancik.
\newblock {LERF}: Language embedded radiance fields.
\newblock In \emph{Proceedings of the IEEE/CVF International Conference on Computer Vision}, pages 19729--19739, 2023.

\bibitem[Khanna et~al.(2024)Khanna, Mao, Jiang, Haresh, Shacklett, Batra, Clegg, Undersander, Chang, and Savva]{khanna2024habitat}
Mukul Khanna, Yongsen Mao, Hanxiao Jiang, Sanjay Haresh, Brennan Shacklett, Dhruv Batra, Alexander Clegg, Eric Undersander, Angel~X Chang, and Manolis Savva.
\newblock Habitat synthetic scenes dataset ({HSSD}-200): An analysis of {3D} scene scale and realism tradeoffs for objectgoal navigation.
\newblock In \emph{Proceedings of the IEEE/CVF Conference on Computer Vision and Pattern Recognition}, pages 16384--16393, 2024.

\bibitem[Kirillov et~al.(2023)Kirillov, Mintun, Ravi, Mao, Rolland, Gustafson, Xiao, Whitehead, Berg, Lo, Doll{\'a}r, and Girshick]{kirillov2023segany}
Alexander Kirillov, Eric Mintun, Nikhila Ravi, Hanzi Mao, Chloe Rolland, Laura Gustafson, Tete Xiao, Spencer Whitehead, Alexander~C. Berg, Wan-Yen Lo, Piotr Doll{\'a}r, and Ross Girshick.
\newblock Segment anything.
\newblock \emph{arXiv:2304.02643}, 2023.

\bibitem[Kuo et~al.(2020)Kuo, Angelova, Lin, and Dai]{kuo2020mask2cad}
Weicheng Kuo, Anelia Angelova, Tsung-Yi Lin, and Angela Dai.
\newblock {Mask2CAD}: {3D} shape prediction by learning to segment and retrieve.
\newblock In \emph{Proceedings of the European Conference on Computer Vision (ECCV)}, pages 260--277. Springer, 2020.

\bibitem[Kuo et~al.(2021)Kuo, Angelova, Lin, and Dai]{kuo2021patch2cad}
Weicheng Kuo, Anelia Angelova, Tsung-Yi Lin, and Angela Dai.
\newblock Patch2cad: Patchwise embedding learning for in-the-wild shape retrieval from a single image.
\newblock In \emph{Proceedings of the IEEE/CVF International Conference on Computer Vision}, pages 12589--12599, 2021.

\bibitem[Langer et~al.(2022)Langer, Bae, Budvytis, and Cipolla]{sparc}
F. Langer, G. Bae, I. Budvytis, and R. Cipolla.
\newblock {SPARC}: Sparse render-and-compare for {CAD} model alignment in a single {RGB} image.
\newblock In \emph{Proc. British Machine Vision Conference}, London, 2022.

\bibitem[Langer et~al.(2024)Langer, Ju, Dikov, Reitmayr, and Ghafoorian]{langer2024fastcad}
Florian Langer, Jihong Ju, Georgi Dikov, Gerhard Reitmayr, and Mohsen Ghafoorian.
\newblock {FastCAD}: Real-time {CAD} retrieval and alignment from scans and videos.
\newblock \emph{arXiv preprint arXiv:2403.15161}, 2024.

\bibitem[Lee et~al.(2024)Lee, Zhang, and Chang]{lee2024duoduo}
Han-Hung Lee, Yiming Zhang, and Angel~X Chang.
\newblock Duoduo {CLIP}: Efficient {3D} understanding with multi-view images.
\newblock \emph{arXiv preprint arXiv:2406.11579}, 2024.

\bibitem[Li et~al.(2024)Li, Hsu, Gu, Pertsch, Mees, Walke, Fu, Lunawat, Sieh, Kirmani, et~al.]{li2024evaluating}
Xuanlin Li, Kyle Hsu, Jiayuan Gu, Karl Pertsch, Oier Mees, Homer~Rich Walke, Chuyuan Fu, Ishikaa Lunawat, Isabel Sieh, Sean Kirmani, et~al.
\newblock Evaluating real-world robot manipulation policies in simulation.
\newblock \emph{arXiv preprint arXiv:2405.05941}, 2024.

\bibitem[Lin and Mu(2024)]{lin2024instructscene}
Chenguo Lin and Yadong Mu.
\newblock {InstructScene}: Instruction-driven {3D} indoor scene synthesis with semantic graph prior.
\newblock \emph{arXiv preprint arXiv:2402.04717}, 2024.

\bibitem[Liu et~al.(2022)Liu, Zheng, Chen, Cui, and Han]{liu2022towards}
Haolin Liu, Yujian Zheng, Guanying Chen, Shuguang Cui, and Xiaoguang Han.
\newblock Towards high-fidelity single-view holistic reconstruction of indoor scenes.
\newblock In \emph{European Conference on Computer Vision}, pages 429--446. Springer, 2022.

\bibitem[Liu et~al.(2023{\natexlab{a}})Liu, Ye, Nie, He, and Han]{liu2023lasa}
Haolin Liu, Chongjie Ye, Yinyu Nie, Yingfan He, and Xiaoguang Han.
\newblock {LASA}: Instance reconstruction from real scans using a large-scale aligned shape annotation dataset, 2023{\natexlab{a}},  \href{https://arxiv.org/abs/2312.12418}{{\ttfamily arXiv:2312.12418}}.

\bibitem[Liu et~al.(2023{\natexlab{b}})Liu, Zhu, Cai, Han, Ling, Porikli, and Su]{liu2023partslip}
Minghua Liu, Yinhao Zhu, Hong Cai, Shizhong Han, Zhan Ling, Fatih Porikli, and Hao Su.
\newblock {PartSLIP}: Low-shot part segmentation for {3D} point clouds via pretrained image-language models.
\newblock In \emph{Proceedings of the IEEE/CVF conference on computer vision and pattern recognition}, pages 21736--21746, 2023{\natexlab{b}}.

\bibitem[Liu et~al.(2024)Liu, Shi, Kuang, Zhu, Li, Han, Cai, Porikli, and Su]{liu2024openshape}
Minghua Liu, Ruoxi Shi, Kaiming Kuang, Yinhao Zhu, Xuanlin Li, Shizhong Han, Hong Cai, Fatih Porikli, and Hao Su.
\newblock {OpenShape}: Scaling up {3D} shape representation towards open-world understanding.
\newblock \emph{Advances in neural information processing systems}, 36, 2024.

\bibitem[Lu et~al.(2023)Lu, Xu, Wei, Xie, Tomizuka, Keutzer, and Zhang]{lu2023open}
Yuheng Lu, Chenfeng Xu, Xiaobao Wei, Xiaodong Xie, Masayoshi Tomizuka, Kurt Keutzer, and Shanghang Zhang.
\newblock Open-vocabulary point-cloud object detection without {3D} annotation.
\newblock In \emph{Proceedings of the IEEE/CVF conference on computer vision and pattern recognition}, pages 1190--1199, 2023.

\bibitem[Maninis et~al.(2022)Maninis, Popov, Nießner, and Ferrari]{maninis2022vid2cad}
Kevis-Kokitsi Maninis, Stefan Popov, Matthias Nießner, and Vittorio Ferrari.
\newblock {Vid2CAD}: {CAD} model alignment using multi-view constraints from videos.
\newblock \emph{IEEE Transactions on Pattern Analysis and Machine Inttelligence}, 2022.

\bibitem[Maninis et~al.(2023)Maninis, Popov, Nie{\ss}ner, and Ferrari]{maninis2023cad}
Kevis-Kokitsi Maninis, Stefan Popov, Matthias Nie{\ss}ner, and Vittorio Ferrari.
\newblock Cad-estate: Large-scale cad model annotation in rgb videos.
\newblock In \emph{Proceedings of the IEEE/CVF International Conference on Computer Vision}, pages 20189--20199, 2023.

\bibitem[Matthias~Minderer(2023)]{minderer2023scaling}
Neil~Houlsby Matthias~Minderer, Alexey~Gritsenko.
\newblock Scaling open-vocabulary object detection.
\newblock \emph{NeurIPS}, 2023.

\bibitem[Nguyen et~al.(2024{\natexlab{a}})Nguyen, Ngo, Kalogerakis, Gan, Tran, Pham, and Nguyen]{nguyen2024open3dis}
Phuc Nguyen, Tuan~Duc Ngo, Evangelos Kalogerakis, Chuang Gan, Anh Tran, Cuong Pham, and Khoi Nguyen.
\newblock {Open3DIS}: Open-vocabulary {3D} instance segmentation with {2D} mask guidance.
\newblock In \emph{Proceedings of the IEEE/CVF Conference on Computer Vision and Pattern Recognition}, pages 4018--4028, 2024{\natexlab{a}}.

\bibitem[Nguyen et~al.(2024{\natexlab{b}})Nguyen, Groueix, Salzmann, and Lepetit]{nguyen2024gigaPose}
Van~Nguyen Nguyen, Thibault Groueix, Mathieu Salzmann, and Vincent Lepetit.
\newblock {GigaPose: Fast and Robust Novel Object Pose Estimation via One Correspondence}.
\newblock In \emph{Proceedings of the IEEE/CVF Conference on Computer Vision and Pattern Recognition}, 2024{\natexlab{b}}.

\bibitem[Nie et~al.(2020)Nie, Han, Guo, Zheng, Chang, and Zhang]{nie2020total3dunderstanding}
Yinyu Nie, Xiaoguang Han, Shihui Guo, Yujian Zheng, Jian Chang, and Jian~Jun Zhang.
\newblock {Total3DUnderstanding}: Joint layout, object pose and mesh reconstruction for indoor scenes from a single image.
\newblock In \emph{Proceedings of the IEEE/CVF Conference on Computer Vision and Pattern Recognition}, pages 55--64, 2020.

\bibitem[Oquab et~al.(2023)Oquab, Darcet, Moutakanni, Vo, Szafraniec, Khalidov, Fernandez, Haziza, Massa, El-Nouby, et~al.]{oquab2023dinov2}
Maxime Oquab, Timoth{\'e}e Darcet, Th{\'e}o Moutakanni, Huy Vo, Marc Szafraniec, Vasil Khalidov, Pierre Fernandez, Daniel Haziza, Francisco Massa, Alaaeldin El-Nouby, et~al.
\newblock {DINO}v2: Learning robust visual features without supervision.
\newblock \emph{arXiv preprint arXiv:2304.07193}, 2023.

\bibitem[Paschalidou et~al.(2021)Paschalidou, Kar, Shugrina, Kreis, Geiger, and Fidler]{paschalidou2021atiss}
Despoina Paschalidou, Amlan Kar, Maria Shugrina, Karsten Kreis, Andreas Geiger, and Sanja Fidler.
\newblock {ATISS}: Autoregressive transformers for indoor scene synthesis.
\newblock \emph{Advances in Neural Information Processing Systems}, 34:\penalty0 12013--12026, 2021.

\bibitem[Peng et~al.(2023)Peng, Genova, Jiang, Tagliasacchi, Pollefeys, Funkhouser, et~al.]{peng2023openscene}
Songyou Peng, Kyle Genova, Chiyu Jiang, Andrea Tagliasacchi, Marc Pollefeys, Thomas Funkhouser, et~al.
\newblock {OpenScene}: {3D} scene understanding with open vocabularies.
\newblock In \emph{Proceedings of the IEEE/CVF conference on computer vision and pattern recognition}, pages 815--824, 2023.

\bibitem[Qin et~al.(2024)Qin, Li, Zhou, Wang, and Pfister]{qin2024langsplat}
Minghan Qin, Wanhua Li, Jiawei Zhou, Haoqian Wang, and Hanspeter Pfister.
\newblock {LangSplat}: {3D} language gaussian splatting.
\newblock In \emph{Proceedings of the IEEE/CVF Conference on Computer Vision and Pattern Recognition}, pages 20051--20060, 2024.

\bibitem[Radford et~al.(2021)Radford, Kim, Hallacy, Ramesh, Goh, Agarwal, Sastry, Askell, Mishkin, Clark, et~al.]{radford2021learning}
Alec Radford, Jong~Wook Kim, Chris Hallacy, Aditya Ramesh, Gabriel Goh, Sandhini Agarwal, Girish Sastry, Amanda Askell, Pamela Mishkin, Jack Clark, et~al.
\newblock Learning transferable visual models from natural language supervision.
\newblock In \emph{International conference on machine learning}, pages 8748--8763. PMLR, 2021.

\bibitem[Ritchie et~al.(2019)Ritchie, Wang, and Lin]{ritchie2019fast}
Daniel Ritchie, Kai Wang, and Yu-an Lin.
\newblock Fast and flexible indoor scene synthesis via deep convolutional generative models.
\newblock In \emph{Proceedings of the IEEE/CVF Conference on Computer Vision and Pattern Recognition}, pages 6182--6190, 2019.

\bibitem[Roberts et~al.(2021)Roberts, Ramapuram, Ranjan, Kumar, Bautista, Paczan, Webb, and Susskind]{roberts2021hypersim}
Mike Roberts, Jason Ramapuram, Anurag Ranjan, Atulit Kumar, Miguel~Angel Bautista, Nathan Paczan, Russ Webb, and Joshua~M Susskind.
\newblock Hypersim: A photorealistic synthetic dataset for holistic indoor scene understanding.
\newblock In \emph{Proceedings of the IEEE/CVF international conference on computer vision}, pages 10912--10922, 2021.

\bibitem[Rozumnyi et~al.(2023)Rozumnyi, Popov, Maninis, Nie{\ss}ner, and Ferrari]{rozumnyi2023estimating}
Denys Rozumnyi, Stefan Popov, Kevis-Kokitsi Maninis, Matthias Nie{\ss}ner, and Vittorio Ferrari.
\newblock Estimating generic {3D} room structures from {2D} annotations.
\newblock In \emph{Advances in Neural Information Processing Systems}, 2023.

\bibitem[Shamos(1978)]{shamos1978computational}
Michael~Ian Shamos.
\newblock \emph{Computational geometry}.
\newblock Yale University, 1978.

\bibitem[Shi et~al.(2023)Shi, Zhi, and Xu]{shi2023planerectr}
Jingjia Shi, Shuaifeng Zhi, and Kai Xu.
\newblock {PlaneRecTR}: Unified query learning for {3D} plane recovery from a single view.
\newblock In \emph{Proceedings of the IEEE/CVF International Conference on Computer Vision}, pages 9377--9386, 2023.

\bibitem[Stekovic et~al.(2020)Stekovic, Hampali, Rad, Sarkar, Fraundorfer, and Lepetit]{stekovic2020general}
Sinisa Stekovic, Shreyas Hampali, Mahdi Rad, Sayan~Deb Sarkar, Friedrich Fraundorfer, and Vincent Lepetit.
\newblock General {3D} room layout from a single view by render-and-compare.
\newblock In \emph{Proceedings of the European Conference on Computer Vision (ECCV)}, pages 187--203. Springer, 2020.

\bibitem[Suvorov et~al.(2022)Suvorov, Logacheva, Mashikhin, Remizova, Ashukha, Silvestrov, Kong, Goka, Park, and Lempitsky]{Suvorov2021ResolutionrobustLM}
Roman Suvorov, Elizaveta Logacheva, Anton Mashikhin, Anastasia Remizova, Arsenii Ashukha, Aleksei Silvestrov, Naejin Kong, Harshith Goka, Kiwoong Park, and Victor~S. Lempitsky.
\newblock Resolution-robust large mask inpainting with {Fourier} convolutions.
\newblock In \emph{Proceedings of the Winter Conference on Applications of Computer Vision (WACV)}, 2022.

\bibitem[Szot et~al.(2021)Szot, Clegg, Undersander, Wijmans, Zhao, Turner, Maestre, Mukadam, Chaplot, Maksymets, et~al.]{szot2021habitat}
Andrew Szot, Alexander Clegg, Eric Undersander, Erik Wijmans, Yili Zhao, John Turner, Noah Maestre, Mustafa Mukadam, Devendra~Singh Chaplot, Oleksandr Maksymets, et~al.
\newblock Habitat 2.0: Training home assistants to rearrange their habitat.
\newblock \emph{Advances in neural information processing systems}, 34:\penalty0 251--266, 2021.

\bibitem[Takmaz et~al.(2023)Takmaz, Fedele, Sumner, Pollefeys, Tombari, and Engelmann]{takmaz2023openmask3d}
Ay{\c{c}}a Takmaz, Elisabetta Fedele, Robert~W Sumner, Marc Pollefeys, Federico Tombari, and Francis Engelmann.
\newblock {OpenMask3D}: Open-vocabulary {3D} instance segmentation.
\newblock \emph{arXiv preprint arXiv:2306.13631}, 2023.

\bibitem[Tam et~al.(2024)Tam, Pun, Wang, Chang, and Savva]{tam2024scenemotifcoder}
Hou In~Ivan Tam, Hou In~Derek Pun, Austin~T Wang, Angel~X Chang, and Manolis Savva.
\newblock {SceneMotifCoder}: Example-driven visual program learning for generating {3D} object arrangements.
\newblock \emph{arXiv preprint arXiv:2408.02211}, 2024.

\bibitem[Tang et~al.(2024)Tang, Nie, Markhasin, Dai, Thies, and Nie{\ss}ner]{tang2024diffuscene}
Jiapeng Tang, Yinyu Nie, Lev Markhasin, Angela Dai, Justus Thies, and Matthias Nie{\ss}ner.
\newblock {DiffuScene}: Denoising diffusion models for generative indoor scene synthesis.
\newblock In \emph{Proceedings of the IEEE/CVF conference on computer vision and pattern recognition}, pages 20507--20518, 2024.

\bibitem[Umeyama(1991)]{umeyama1991least}
Shinji Umeyama.
\newblock Least-squares estimation of transformation parameters between two point patterns.
\newblock \emph{IEEE Transactions on Pattern Analysis \& Machine Intelligence}, 13\penalty0 (04):\penalty0 376--380, 1991.

\bibitem[Wang et~al.(2018)Wang, Savva, Chang, and Ritchie]{wang2018deep}
Kai Wang, Manolis Savva, Angel~X Chang, and Daniel Ritchie.
\newblock Deep convolutional priors for indoor scene synthesis.
\newblock \emph{ACM Transactions on Graphics (TOG)}, 37\penalty0 (4):\penalty0 1--14, 2018.

\bibitem[Wang et~al.(2019)Wang, Lin, Weissmann, Savva, Chang, and Ritchie]{wang2019planit}
Kai Wang, Yu-An Lin, Ben Weissmann, Manolis Savva, Angel~X Chang, and Daniel Ritchie.
\newblock {PlanIT}: Planning and instantiating indoor scenes with relation graph and spatial prior networks.
\newblock \emph{ACM Transactions on Graphics (TOG)}, 38\penalty0 (4):\penalty0 1--15, 2019.

\bibitem[Wang et~al.(2024)Wang, Fan, Wang, Su, Ramamoorthi, et~al.]{wang2024lift3d}
Peihao Wang, Zhiwen Fan, Zhangyang Wang, Hao Su, Ravi Ramamoorthi, et~al.
\newblock {Lift3D}: Zero-shot lifting of any {2D} vision model to {3D}.
\newblock In \emph{Proceedings of the IEEE/CVF Conference on Computer Vision and Pattern Recognition}, pages 21367--21377, 2024.

\bibitem[Wang et~al.(2021)Wang, Yeshwanth, and Nie{\ss}ner]{wang2021sceneformer}
Xinpeng Wang, Chandan Yeshwanth, and Matthias Nie{\ss}ner.
\newblock {SceneFormer}: Indoor scene generation with transformers.
\newblock In \emph{2021 International Conference on 3D Vision (3DV)}, pages 106--115. IEEE, 2021.

\bibitem[Wei et~al.(2023)Wei, Ding, Park, Sajnani, Poulenard, Sridhar, and Guibas]{wei2023lego}
Qiuhong~Anna Wei, Sijie Ding, Jeong~Joon Park, Rahul Sajnani, Adrien Poulenard, Srinath Sridhar, and Leonidas Guibas.
\newblock Lego-net: Learning regular rearrangements of objects in rooms.
\newblock In \emph{Proceedings of the IEEE/CVF Conference on Computer Vision and Pattern Recognition}, pages 19037--19047, 2023.

\bibitem[Wu et~al.(2024{\natexlab{a}})Wu, Raychaudhuri, Ritchie, Savva, and Chang]{wu2024r3ds}
Qirui Wu, Sonia Raychaudhuri, Daniel Ritchie, Manolis Savva, and Angel~X Chang.
\newblock R3ds: Reality-linked 3d scenes for panoramic scene understanding.
\newblock In \emph{Proceedings of the European Conference on Computer Vision (ECCV)}, 2024{\natexlab{a}}.

\bibitem[Wu et~al.(2024{\natexlab{b}})Wu, Ritchie, Savva, and Chang]{wu2024generalizing}
Qirui Wu, Daniel Ritchie, Manolis Savva, and Angel~X Chang.
\newblock Generalizing single-view {3D} shape retrieval to occlusions and unseen objects.
\newblock In \emph{2024 International Conference on 3D Vision (3DV)}, pages 893--902. IEEE, 2024{\natexlab{b}}.

\bibitem[Xue et~al.(2024)Xue, Yu, Zhang, Panagopoulou, Li, Mart{\'\i}n-Mart{\'\i}n, Wu, Xiong, Xu, Niebles, et~al.]{xue2024ulip}
Le Xue, Ning Yu, Shu Zhang, Artemis Panagopoulou, Junnan Li, Roberto Mart{\'\i}n-Mart{\'\i}n, Jiajun Wu, Caiming Xiong, Ran Xu, Juan~Carlos Niebles, et~al.
\newblock {ULIP}-2: Towards scalable multimodal pre-training for {3D} understanding.
\newblock In \emph{Proceedings of the IEEE/CVF Conference on Computer Vision and Pattern Recognition}, pages 27091--27101, 2024.

\bibitem[Yang et~al.(2022)Yang, Zheng, Dai, Tang, Ma, and Yuan]{Yang_2022_WACV}
Cheng Yang, Jia Zheng, Xili Dai, Rui Tang, Yi Ma, and Xiaojun Yuan.
\newblock Learning to reconstruct 3d non-cuboid room layout from a single rgb image.
\newblock In \emph{Proceedings of the IEEE/CVF Winter Conference on Applications of Computer Vision (WACV)}, 2022.

\bibitem[Yang et~al.(2023)Yang, Zhang, Li, Zou, Li, and Gao]{yang2023setofmark}
Jianwei Yang, Hao Zhang, Feng Li, Xueyan Zou, Chunyuan Li, and Jianfeng Gao.
\newblock Set-of-mark prompting unleashes extraordinary visual grounding in {GPT-4V}.
\newblock \emph{arXiv preprint arXiv:2310.11441}, 2023.

\bibitem[Yang et~al.(2024{\natexlab{a}})Yang, Kang, Huang, Zhao, Xu, Feng, and Zhao]{depth_anything_v2}
Lihe Yang, Bingyi Kang, Zilong Huang, Zhen Zhao, Xiaogang Xu, Jiashi Feng, and Hengshuang Zhao.
\newblock Depth anything v2.
\newblock \emph{arXiv:2406.09414}, 2024{\natexlab{a}}.

\bibitem[Yang et~al.(2024{\natexlab{b}})Yang, Ju, and Yi]{yang2024imov3d}
Timing Yang, Yuanliang Ju, and Li Yi.
\newblock {ImOV3D}: Learning open-vocabulary point clouds {3D} object detection from only {2D} images.
\newblock \emph{arXiv preprint arXiv:2410.24001}, 2024{\natexlab{b}}.

\bibitem[Yang et~al.(2024{\natexlab{c}})Yang, Sun, Weihs, VanderBilt, Herrasti, Han, Wu, Haber, Krishna, Liu, et~al.]{yang2024holodeck}
Yue Yang, Fan-Yun Sun, Luca Weihs, Eli VanderBilt, Alvaro Herrasti, Winson Han, Jiajun Wu, Nick Haber, Ranjay Krishna, Lingjie Liu, et~al.
\newblock Holodeck: Language guided generation of {3D} embodied {AI} environments.
\newblock In \emph{Proceedings of the IEEE/CVF Conference on Computer Vision and Pattern Recognition}, pages 16227--16237, 2024{\natexlab{c}}.

\bibitem[Yu et~al.(2024)Yu, Zhao, Pang, Zhang, and Lu]{yu2024multi}
Qian Yu, Xiaoqi Zhao, Youwei Pang, Lihe Zhang, and Huchuan Lu.
\newblock Multi-view aggregation network for dichotomous image segmentation.
\newblock \emph{arXiv:2404.07445}, 2024.

\bibitem[Yu et~al.(2023)Yu, Feng, Feng, Liu, Jin, Zeng, and Chen]{yu2023inpaint}
Tao Yu, Runseng Feng, Ruoyu Feng, Jinming Liu, Xin Jin, Wenjun Zeng, and Zhibo Chen.
\newblock Inpaint anything: Segment anything meets image inpainting.
\newblock \emph{arXiv:2304.06790}, 2023.

\bibitem[Yue et~al.(2025)Yue, Das, Engelmann, Tang, and Lenssen]{yue2025improving}
Yuanwen Yue, Anurag Das, Francis Engelmann, Siyu Tang, and Jan~Eric Lenssen.
\newblock Improving {2D} feature representations by {3D}-aware fine-tuning.
\newblock In \emph{European Conference on Computer Vision}, pages 57--74. Springer, 2025.

\bibitem[Zhang et~al.(2021)Zhang, Cui, Chen, Liu, Zeng, Bao, and Zhang]{zhang2021deeppanocontext}
Cheng Zhang, Zhaopeng Cui, Cai Chen, Shuaicheng Liu, Bing Zeng, Hujun Bao, and Yinda Zhang.
\newblock {DeepPanoContext}: Panoramic {3D} scene understanding with holistic scene context graph and relation-based optimization.
\newblock In \emph{Proceedings of the IEEE/CVF International Conference on Computer Vision}, pages 12632--12641, 2021.

\bibitem[Zhang et~al.(2023)Zhang, Dong, and Ma]{zhang2023clip}
Junbo Zhang, Runpei Dong, and Kaisheng Ma.
\newblock {CLIP-FO3D}: Learning free open-world {3D} scene representations from {2D} dense {CLIP}.
\newblock In \emph{Proceedings of the IEEE/CVF International Conference on Computer Vision}, pages 2048--2059, 2023.

\bibitem[Zheng et~al.(2020)Zheng, Zhang, Li, Tang, Gao, and Zhou]{zheng2020structured3d}
Jia Zheng, Junfei Zhang, Jing Li, Rui Tang, Shenghua Gao, and Zihan Zhou.
\newblock {Structured3D}: A large photo-realistic dataset for structured {3D} modeling.
\newblock In \emph{Proceedings of the European Conference on Computer Vision (ECCV)}, pages 519--535. Springer, 2020.

\bibitem[Zheng et~al.(2024)Zheng, Gao, Fan, Liu, Laaksonen, Ouyang, and Sebe]{zheng2024birefnet}
Peng Zheng, Dehong Gao, Deng-Ping Fan, Li Liu, Jorma Laaksonen, Wanli Ouyang, and Nicu Sebe.
\newblock Bilateral reference for high-resolution dichotomous image segmentation.
\newblock \emph{CAAI Artificial Intelligence Research}, 3:\penalty0 9150038, 2024.

\bibitem[Zhou et~al.(2024)Zhou, Liu, and Han]{zhou2024zero}
Junsheng Zhou, Yu-Shen Liu, and Zhizhong Han.
\newblock Zero-shot scene reconstruction from single images with deep prior assembly.
\newblock \emph{arXiv preprint arXiv:2410.15971}, 2024.

\bibitem[Zhou et~al.(2018)Zhou, Park, and Koltun]{Zhou2018open3d}
Qian-Yi Zhou, Jaesik Park, and Vladlen Koltun.
\newblock {Open3D}: {A} modern library for {3D} data processing.
\newblock \emph{arXiv:1801.09847}, 2018.

\bibitem[Zhu et~al.(2023)Zhu, Zhang, He, Guo, Zeng, Qin, Zhang, and Gao]{zhu2023pointclip}
Xiangyang Zhu, Renrui Zhang, Bowei He, Ziyu Guo, Ziyao Zeng, Zipeng Qin, Shanghang Zhang, and Peng Gao.
\newblock {PointCLIP} v2: Prompting {CLIP} and {GPT} for powerful {3D} open-world learning.
\newblock In \emph{Proceedings of the IEEE/CVF International Conference on Computer Vision}, pages 2639--2650, 2023.

\bibitem[Zook et~al.(2024)Zook, Sun, Spjut, Blukis, Birchfield, and Tremblay]{zook2024grs}
Alex Zook, Fan-Yun Sun, Josef Spjut, Valts Blukis, Stan Birchfield, and Jonathan Tremblay.
\newblock {GRS}: Generating robotic simulation tasks from real-world images.
\newblock \emph{arXiv preprint arXiv:2410.15536}, 2024.

\bibitem[Zou et~al.(2023)Zou, Yang, Zhang, Li, Li, Wang, Wang, Gao, and Lee]{zou2023segment}
Xueyan Zou, Jianwei Yang, Hao Zhang, Feng Li, Linjie Li, Jianfeng Wang, Lijuan Wang, Jianfeng Gao, and Yong~Jae Lee.
\newblock Segment everything everywhere all at once.
\newblock In \emph{Advances in Neural Information Processing Systems}, 2023.

\end{thebibliography}
}

\clearpage
\appendix

\begin{figure*}[t]
\centering
\includegraphics[width=\linewidth]{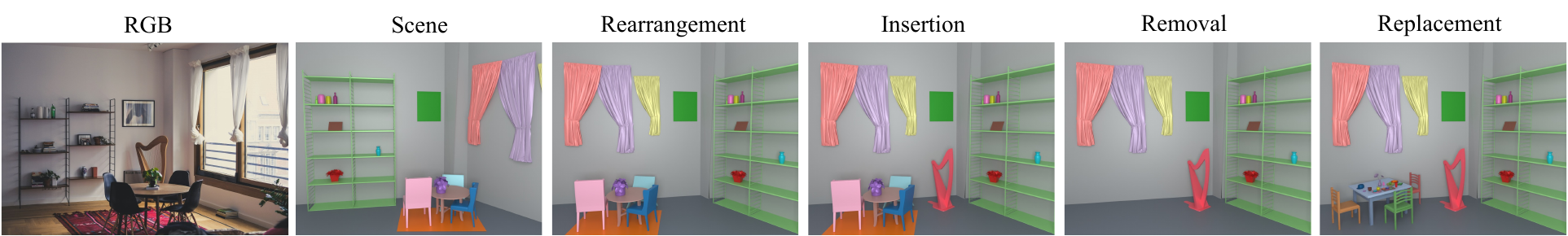}
\caption{
We show the potential of Diorama in flexible scene editing in terms of different editing types, including object rearrangement, insertion, removal and replacement.
} 
\label{fig:editing}
\end{figure*}

In this supplement, we provide additional details of our method (\Cref{sec:supp-method}) and experiments (\Cref{sec:supp-experiments}). In \Cref{sec:supp-examples}, we present more qualitative results of Diorama on SSDB images (\cref{fig:wss_results_supp}), as well as on real-world internet images (\cref{fig:real_supp}) and text-conditioned generated images (\cref{fig:text2scene_supp}). Particularly, we demonstrate the potential application of Diorama in flexible scene editing in~\cref{fig:editing}.

\section{Method details}
\label{sec:supp-method}

For all experiments involving LMM-based visual reasoning, we use the checkpoint ``gpt-4o-2024-08-06'' of GPT-4o. It costs approximately \$0.12 on average per SSDB image for 344 examples.

\subsection{Holistic scene parsing}
\label{sec:supp-scene-parsing}

Holistic scene parsing consists of several tasks to comprehensively understanding the semantics and geometry of a scene image that serve as inputs to downstream components, including object recognition and localization, and depth and normal estimation.

For open-world object recognition, given an image, we aim to identify the objects in the scene by prompting GPT-4o and run open-vocabulary detectors and segmentation models to obtain the bounding boxes and masks. Below, we provide prompts and additional details.

Prompt for identifying objects in an image:
\begin{minted}[breaklines,
breaksymbol={},
fontsize=\scriptsize]{text}
What are objects in the image? Includes architectural elements such as floor and wall if applicable. Ensure that each class is singular and has no quantifiers.

Return the output in a JSON format according to the following format:
{
  "classes": [object1, object2, object3, ...]
}
\end{minted}

To obtain object bounding boxes and masks for localization, we first run the detector OWLv2~\cite{minderer2023scaling} by providing text inputs of a template prompt \textit{``a photo of CLASS''} and obtain detection results after non-maximal suppression. 
To ensure each object instance is only captured by one detection box to avoid repeated 3D shape retrieval for the same instance, we apply class-aware multiple-instance suppression where the bounding box of the same category with the largest intersection-over-self (IoS) is discarded. We then prompt SAM~\cite{kirillov2023segany} with each detected bounding box to predict the segmentation mask for each object by assigning the one with the highest score.

\subsection{LMM-powered scene graph generation}
\label{sec:supp-scene-graph}

\begin{figure}[t]
\centering
\includegraphics[width=\linewidth]{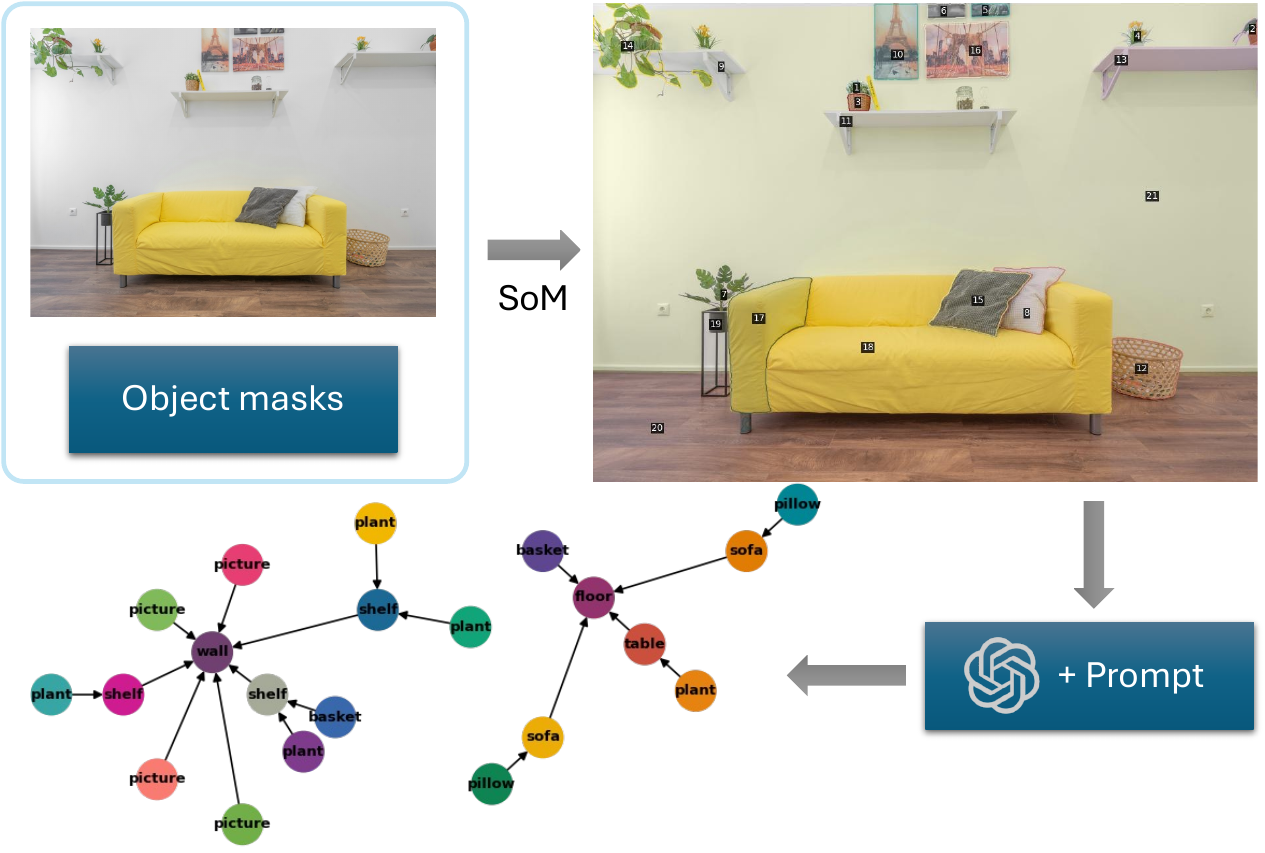}
\caption{
Illustration of scene graph generation using GPT-4o.
} 
\label{fig:llm-grounding}
\end{figure}

After obtaining the set of objects in the scene, we then use an LMM (e.g. GPT-4o) to obtain the scene-graph. The original input image is augmented by partitioning into semantically meaningful regions and overlaying a set of visual marks on it.
Specifically, we use predicted object masks $\mathbf{M}$ to represent image regions of interest and each region is marked by its corresponding $\langle\textit{object id}\rangle$ that can be recognized and referred by GPT-4o using its OCR and visual reasoning capabilities. The generated scene-graph provide information about relationships between objects, including support relations.

We specify to the LMM the desired output response in JSON format so it can be successfully executed by the program afterward. We also limit object supporting relations to be selected from [``placed on'', ``mounted on''].
To encourage the LMM to obey the provided numeric object marks and reduce hallucination errors, we use interleaved text prompt by incorporating the object marks and template response format into the scene graph generation prompt directly for symbolic reference.

Prompt for extracting a scene graph for objects in a image.  Note the objects were already identified and provided as input.
\begin{minted}[breaklines,
breaksymbol={},
fontsize=\scriptsize]{text}
OBJECT_ENTRY = {{
  "id": {obj_id} (id of object),
  "name": {obj_name} (name of object),
  "caption": brief description of object appearance (excluding relationship),
  "face_direction": direction of the object (choose from the list {face_direction}),
}},

SUPPORT_RELATION_ENTRY = {{
  "subject_id": {obj_id} (id of object),
  "type": choose from a list of supporting relationships {support_rel},
  "target_id": id of supporting object,
    }},

Please follow the examples in the Visual Genome dataset and generate a scene graph that best describes the image. A reference list of labelled objects is provided as "{objects_with_ids}" in the format of <object_id, object_name>.
Return the output in JSON according to the following format:
{{
  "room_type": type of room,
  "objects": [
    {object_list}
  ],
  "support_relation": [
    {support_relation_list}
  ]
}}

Relationships are expressed as <subject_id, type, target_id>. Each pair of <subject_id, target_id> only occurs once.
\end{minted}

\subsection{Architecture reconstruction}
\label{sec:supp-architecture}

To obtain a planar reconstruction of the architecture, we apply a three-step process: 1) segmenting out objects and inpainting to obtain empty rooms, 2) using depth estimation to obtain 3D points, and 3) clustering points by normals and plane fitting to obtain planar architecture surfaces. 

\mypara{Segmentation and inpainting to obtain empty room.} We begin by obtaining object masks that are used to inpaint the images, leaving us with unfurnished scenes.
For SSDB, we use dichotomous segmentation methods such as BiRefNet~\cite{zheng2024birefnet} and MVANet~\cite{yu2024multi}. To maximize the recall of predicted masks, we apply each method separately twice by filling the mask predicted in the first iteration with white color then re-running the models on such images and merging the masks from both iterations. 

We find that BiRefNet excels at producing sharp and complete masks for large objects and dense object arrangements, while MVANet has higher recall for scenes without a single cluster of objects.
Therefore, we opt to merge the masks produced by these two methods.
Additionally, we use SEEM~\cite{zou2023segment} for all other experiments as we find that scenes featuring objects heavily dispersed around the scene are challenging for dichotomous segmentation networks. We note that the benefit of using dichotomous segmentation compared to other segmentation networks is in completeness of the masks (e.g., they cover the full area of the object with higher probability). Ensuring that the masks are as complete as possible is crucial for the further steps of PlainRecon.

After obtaining a segmentation of the objects, we `erase' the objects by applying inpainting to defurnish the room.
We use LaMa~\cite{Suvorov2021ResolutionrobustLM} to inpaint the background (similar to  \citet{yu2023inpaint}).
While recent advances in generative models are producing higher-quality inpainting, we find that even specialized inpainting for removal methods such as CLIPAway ~\cite{ekin2024clipaway} still suffer from hallucinating rather than inpainting empty background.
This step should ideally produce perfectly defurnished and sharp image.
In practice, the resulting image is not fully defurnished and the inpainted region is blurry. 
While the former remains a bottleneck of the pipeline, we notice that the latter is frequently alleviated by robustness of monocular depth estimation models which is the next step of the pipeline.

\mypara{Depth estimation to obtain point cloud.} After we have a defurnished room, we apply depth estimation to obtain a 3D point cloud of the scene.
We compare two depth estimation models applied on the inpainted images - DepthAnythingV2 (DAv2) ~\cite{depth_anything_v2} and Metric3D (M3D) ~\cite{hu2024metric3d}. 
As we require normals for clustering points into planes, we leverage M3D normal estimation capabilities. 
We note that the predicted normals are more robust to the blurriness of the inpainting step compared to depth. 
We project the pixels to point clouds using the predicted metric depth. 
At the end of this process, we have point clouds with normals of unfurnished scenes, where the variation in normals should now be exclusively explained by presence of different planes in scene architecture. 

\mypara{Plane identification and fitting.} 
The next step is normals-based clustering of the points into planar segments. 
We begin by pre-processing point clouds by applying voxel downsampling and removing statistical outliers with both using implementations from Open3D~\cite{Zhou2018open3d}. Since voxel downsampling bins the points in each voxel into one, the resulting point cloud has an approximately even distance between pairs of neighboring points.
This ensures stability of hyperparameters in the subsequent steps and filters some noise introduced by depth estimation models.
Next we run K-means clustering ($k_m=12$ to account for noise) on normals to determine seed normals. 
We iteratively select the seed with the largest number of corresponding points, and cluster all the points that have normals within the angle threshold ($\alpha = 10$ in our experiments).
We further apply DBScan ~\cite{ester1996dbscan} to separate the initial cluster as it is possible to have multiple walls with identical normals in the scene. 
The algorithm terminates when we either run out of seed normals or have less than the threshold number of unclustered points left ($N_{\min} = 200$). 
Finally, we propagate the instance labels to full point clouds using KNN ($k_{n}=1$). 
We assign the $floor$ label to the largest cluster with a normal pointing sufficiently upwards.
Similarly, we filter out ceiling clusters as those points with a downward facing normal.

Once we have obtained the point clusters, we fit bounded planar segments to the points to obtain the final architecture.
We start by fitting a plane using RANSAC~\cite{fischler1981ransac} to obtain a plane equation, and then estimate the bounding boxes of architecture elements. 
To obtain tight boxes, we compute convex hulls of point cloud segments, project the convex hull vertices onto 2D plane and apply the rotating calipers algorithm ~\cite{shamos1978computational} to obtain the bounding box with the smallest area. 
Finally, we convert vertices back into 3D. While such initialization provides good plane bounds, we need to refine the plane further to account for imperfections of previous steps. 
We begin by making sure that all the walls are orthogonal to the floor, this is simply done by solving a system of linear equations that finds a vector orthogonal to the intersection line of floor and wall and lies on the floor plane. 
We proceed to use this vector as a new normal for our wall equation. Then we proceed to refine the plane bounds by finding the intersections of each architecture element, projecting two closest vertices of one plane onto the intersection line and adjusting the vertices of the plane we projected onto to the projected vertices of the other plane, effectively connecting them. Finally, we duplicate the vertices with a small offset along the direction opposite to normal and export the meshes of architectural components.

\begin{table}%
\centering
\resizebox{\linewidth}{!}
{
\begin{tabular}{@{}l cccccc @{}}
\toprule
\multirow{2}{*}{Method}  & \multicolumn{4}{c}{Scene-aware Alignment $\uparrow$} & \multirow{2}{*}{Collision $\downarrow$} & \multirow{2}{*}{Relation $\uparrow$} \\
\cmidrule(lr){2-5} 
& rAcc & tAcc & sAcc & Acc &  &   \\
\midrule
BM baseline & 0.34 & 0.84 & 0.44 & 0.15 & 12.29 & 0.58  \\
ZSP~\cite{goodwin2022zero} & 0.32 & 0.88 & 0.50 & 0.18 & 8.87 & 0.59  \\
ZSP w/ DinoV2 & 0.33 & 0.81 & 0.48 & 0.20 & 9.80 & 0.57 \\
ZSP w/ ft DinoV2 & 0.34 & 0.83 & 0.50 & 0.21 & 9.97 & 0.59  \\
GigaPose~\cite{nguyen2024gigaPose} & 0.36 & 0.89 & \textbf{0.67} & 0.24 & 7.43 & 0.61 \\
Ours & \textbf{0.41} & \textbf{0.91} & 0.61 & \textbf{0.28} & \textbf{6.93}  & \textbf{0.61} \\
\bottomrule
\end{tabular}
}
\vspace{-6pt}
\caption{
Additional comparison of zero-shot pose estimation methods for the 9-DoF CAD alignment task on SSDB images using estimated depth by Metric3DV2.
}
\label{tab:wss_pose_predd}
\end{table}

\begin{table}[t]
\centering
\resizebox{\linewidth}{!}
{
\begin{tabular}{@{}l c cccc @{}}
\toprule
Groups & \#Cats / \#Insts & rAcc $\uparrow$ & tAcc $\uparrow$ & sAcc $\uparrow$ & Acc $\uparrow$ \\
\midrule
Household & 466 / 5577 & 0.46 & \textbf{0.87} & \textbf{0.71} & 0.32 \\
Furniture & 24 / 1137 & \textbf{0.54} & 0.86 & \textbf{0.71} & \textbf{0.39} \\
\midrule
Occluded & 369 / 3175 & 0.44 & 0.84 & 0.62 & 0.28 \\
Complete & 401 / 3539 & \textbf{0.50} & \textbf{0.89} & \textbf{0.78} & \textbf{0.38} \\
\midrule
Supported & 468 / 5715 & \textbf{0.48} & \textbf{0.87} & \textbf{0.71} & \textbf{0.34} \\
Supporting & 79 / 1634 & 0.46 & 0.84 & 0.69 & 0.31  \\
\bottomrule
\end{tabular}
}
\caption{Averaged alignment results across different SSDB object groups.
For each group, we count the number of fine-grained categories and object instances.
Occluded objects are those with occlusion ratio above a threshold 5\% of pixels.
}
\label{tab:wss_groups}
\end{table}

\begin{table}[t]
\centering
\resizebox{\linewidth}{!}
{
\begin{tabular}{@{}c cccc cccc @{}}
\toprule
\multirow{2}{*}{\# retrievals} & \multicolumn{4}{c}{Ground Truth Depth} &\multicolumn{4}{c}{Estimated Depth} \\
\cmidrule(lr){2-5} \cmidrule(lr){6-9} 
 & rAcc & tAcc & sAcc & Acc & rAcc & tAcc & sAcc & Acc \\
\midrule
1 & 0.22 & 0.92 & 0.52 & 0.11 & 0.21 & 0.88 & 0.47 & 0.10  \\
4 & 0.32 & 0.96 & 0.68 & 0.20 & 0.32 & 0.93 & 0.63 & 0.18 \\
8 & \textbf{0.37} & \textbf{0.97} & \textbf{0.72} & \textbf{0.23} & \textbf{0.37} & \textbf{0.94} & \textbf{0.68} & \textbf{0.21} \\
\bottomrule
\end{tabular}
}
\caption{9D CAD alignment results given a different number of retrieved 3D shapes using either ground-truth or estimated depth.
}
\label{tab:wss_hypo}
\end{table}

\begin{table}[t]
\centering
\resizebox{\linewidth}{!}
{
\begin{tabular}{@{}l c ccc @{}}
\toprule
\multirow{2}{*}{Ablation} & \multirow{2}{*}{Collision $\downarrow$} & \multicolumn{3}{c}{Scene Structure $\uparrow$}  \\
\cmidrule(lr){3-5} 
 &  & orientation & placement & overall \\
\midrule
all-in-one & 12.80 & 0.19 & 0.92 & 0.16 \\
stage-wise & \textbf{3.78} & \textbf{0.98} & \textbf{0.95} & \textbf{0.93} \\
\bottomrule
\end{tabular}
}
\caption{Comparison between all-in-one and stage-wise optimization.
}
\label{tab:wss_lo_add}
\end{table}

\begin{table*}
\centering
\resizebox{\linewidth}{!}
{
\begin{tabular}{@{}l cc cccccccccccc @{}}
\toprule
Method & supervision & \#hypo & bed & bkshlf & cabinet & chair & sofa & table & bin & bathtub & display & {\color{red}others} & cls. avg & ist. avg \\
\midrule
ROCA & \cmark\cmark & - & 7.02 & 3.62 & 7.56 & 20.03 & 5.26 & 9.16 & 13.19 & 8.11 & 13.21 & {\color{lightgray}0.00*} & 8.72 & 12.87 \\
\midrule
DiffCAD & \cmark & 1 & 7.02 & 0.00 & 3.36 & 9.38 & 6.58 & 1.58 & {\color{lightgray}0.00*} & {\color{lightgray}0.00*} & {\color{lightgray}0.00*} & {\color{lightgray}0.00*} & 2.79 & 4.59\\
DiffCAD & \cmark & 5 & 12.28 & 0.72 & 6.72 & 14.2 & 7.89 & 1.58 & {\color{lightgray}0.00*} & {\color{lightgray}0.00*} & {\color{lightgray}0.00*} & {\color{lightgray}0.00*} & 4.34 & 6.84 \\
DiffCAD & \cmark & 10 & 7.02 & 1.45 & 4.21 & 11.41 & 5.26 & 1.26 & {\color{lightgray}0.00*} & {\color{lightgray}0.00*} & {\color{lightgray}0.00*} & {\color{lightgray}0.00*} & 3.06 & 5.41 \\
\midrule
Ours & \xmark & - & 0.00 & 0.74 & 4.24 & 9.72 & 3.95 & 4.27 & 0.00 & 0.00 & 10.36 & 5.56 & 3.33 & 6.66 \\
\bottomrule
\end{tabular}
}
\caption{Object-focus alignment accuracy on the Scan2CAD benchmark. ROCA is fully-supervised using in-domain data. DiffCAD is weakly-supervised using synthetic data.}
\label{tab:s2c_results}
\end{table*}

\begin{figure}%
\centering
\includegraphics[width=\linewidth]{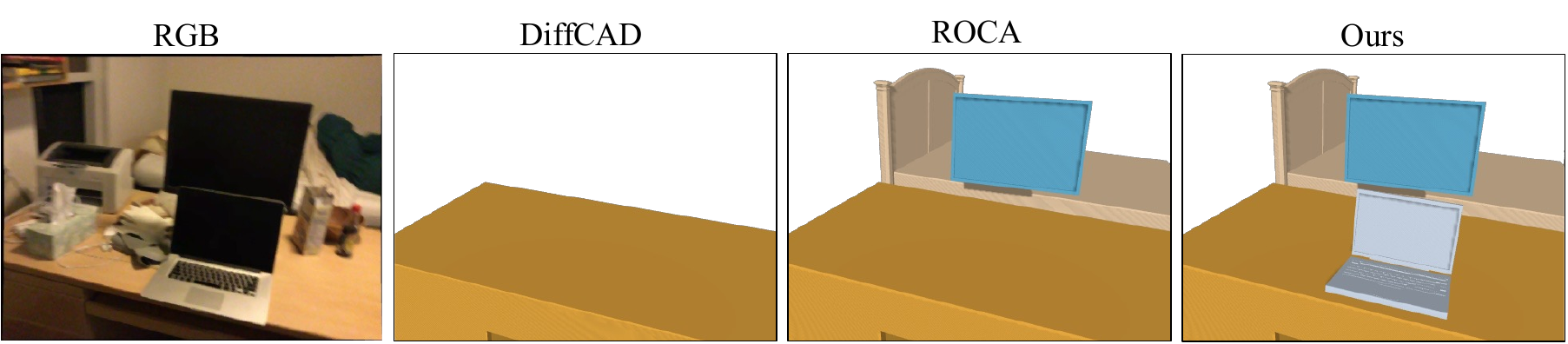}
\caption{
Qualitative comparison between different ground-truth evaluation sets used in DiffCAD, ROCA and ours.
} 
\vspace{-6pt}
\label{fig:s2c_eval_diff}
\end{figure}

\begin{figure*}[t]
\centering
\includegraphics[width=\linewidth]{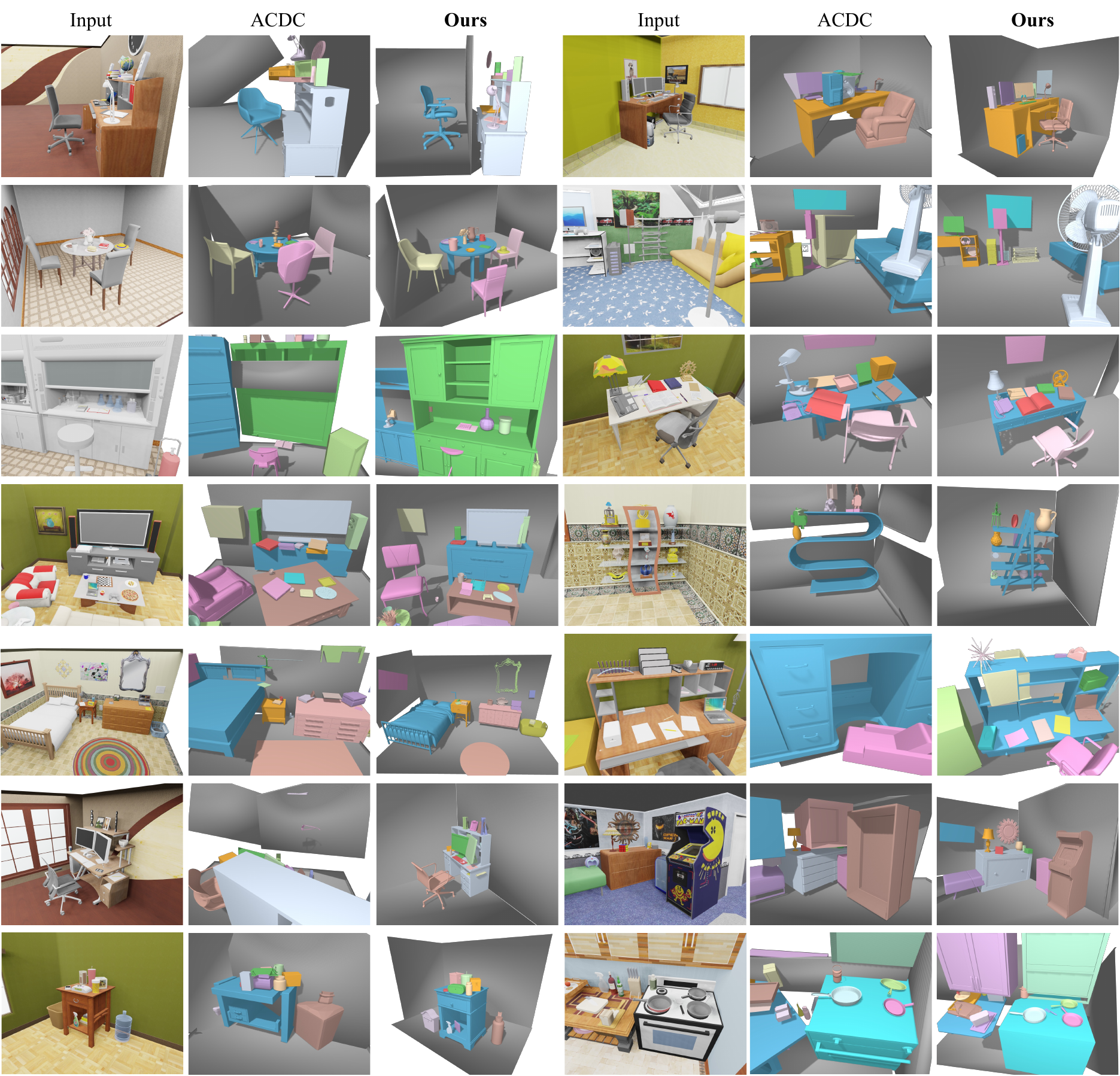}
\caption{
More comparison examples on SSDB.
}
\label{fig:main_results_supp}
\end{figure*}

\begin{figure*}[t]
\centering
\includegraphics[width=\linewidth]{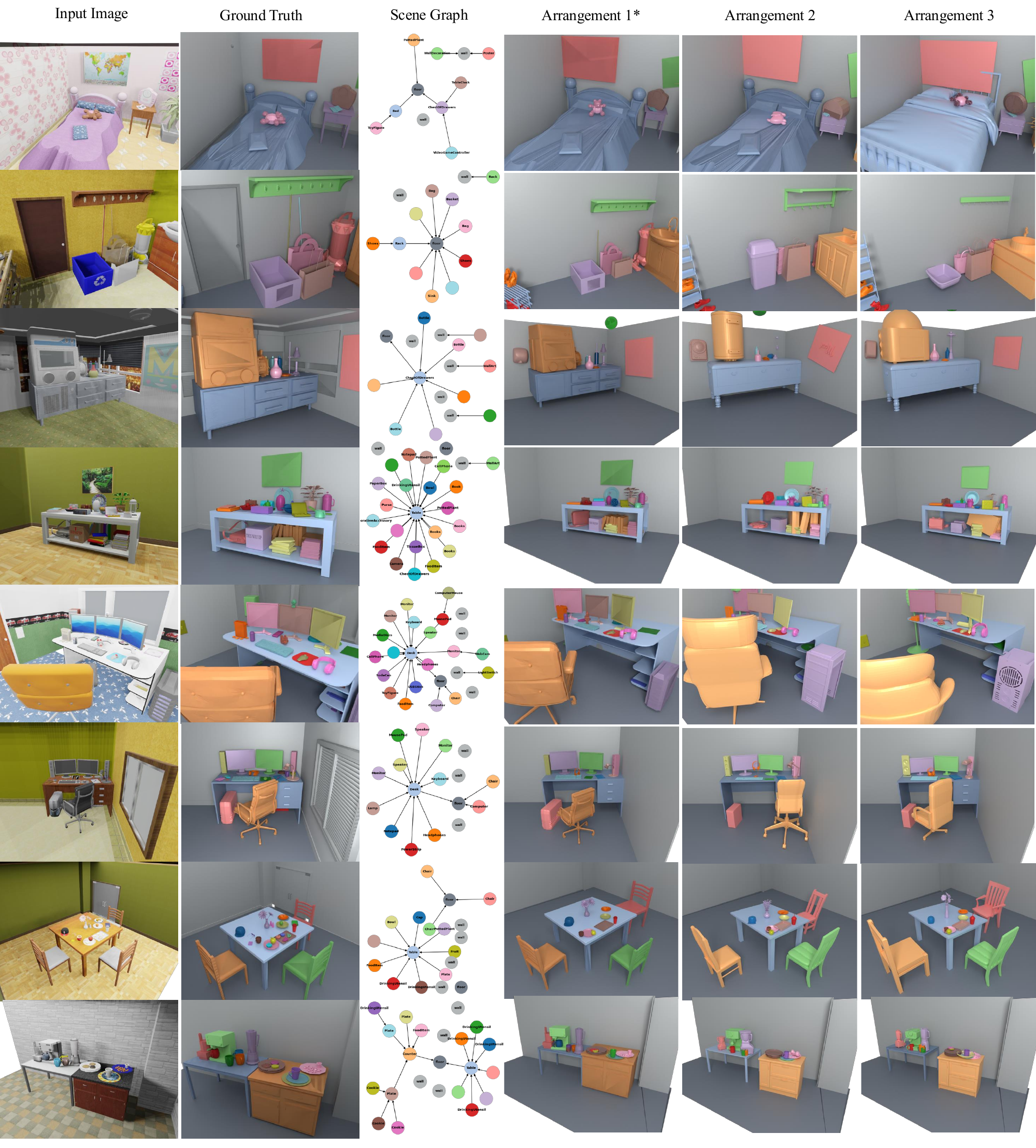}
\caption{
More Diorama examples on SSDB images. 
}
\label{fig:wss_results_supp}
\end{figure*}

\begin{figure*}%
\centering
\setkeys{Gin}{width=\linewidth}
\setlength{\tabcolsep}{3pt}
\begin{tabularx}{\linewidth}{@{} XXXX @{}}
\multicolumn{1}{c}{Input} & \multicolumn{1}{c}{Output} & \multicolumn{1}{c}{Input} & \multicolumn{1}{c}{Output} \\
\imgclip{0}{fig/images/wild_figure_supp/scene_03_gt} & \imgclip{0}{fig/images/wild_figure_supp/scene_03_diorama} & \imgclip{0}{fig/images/wild_figure_supp/scene_04_gt} & \imgclip{0}{fig/images/wild_figure_supp/scene_04_diorama}  \\
\imgclip{0}{fig/images/wild_figure_supp/scene_11_gt} & \imgclip{0}{fig/images/wild_figure_supp/scene_11_diorama} & \imgclip{0}{fig/images/wild_figure/scene_00_gt} & \imgclip{0}{fig/images/wild_figure/scene_00_diorama}  \\
\imgclip{0}{fig/images/wild_figure_supp/scene_21_gt} & \imgclip{0}{fig/images/wild_figure_supp/scene_21_diorama} & \imgclip{0}{fig/images/wild_figure_supp/scene_23_gt} & \imgclip{0}{fig/images/wild_figure_supp/scene_23_diorama}  \\
\end{tabularx}
\caption{
More examples for real-world internet images. 
}
\label{fig:real_supp}
\end{figure*}

\begin{figure*}%
\centering
\setkeys{Gin}{width=\linewidth}
\setlength{\tabcolsep}{3pt}
\begin{tabularx}{\linewidth}{@{} XXX XXX @{}}
\multicolumn{1}{c}{Text prompt} & \multicolumn{1}{c}{Generated image} & \multicolumn{1}{c}{Output} & \multicolumn{1}{c}{Text prompt} & \multicolumn{1}{c}{Generated image} & \multicolumn{1}{c}{Output} \\
\imgclip{0}{fig/images/flux_figure/scene_10_prompt} & \imgclip{0}{fig/images/flux_figure/scene_10_gt} & \imgclip{0}{fig/images/flux_figure/scene_10_diorama} & \imgclip{0}{fig/images/flux_figure_supp/scene_23_prompt} &  \imgclip{0}{fig/images/flux_figure_supp/scene_23_gt} & \imgclip{0}{fig/images/flux_figure_supp/scene_23_diorama}  \\
\imgclip{0}{fig/images/flux_figure_supp/scene_29_prompt}& \imgclip{0}{fig/images/flux_figure_supp/scene_29_gt} & \imgclip{0}{fig/images/flux_figure_supp/scene_29_diorama} & \imgclip{0}{fig/images/flux_figure_supp/scene_40_prompt} & \imgclip{0}{fig/images/flux_figure_supp/scene_40_gt} & \imgclip{0}{fig/images/flux_figure_supp/scene_40_diorama}  \\
\imgclip{0}{fig/images/flux_figure_supp/scene_62_prompt}& \imgclip{0}{fig/images/flux_figure_supp/scene_62_gt} & \imgclip{0}{fig/images/flux_figure_supp/scene_62_diorama} & \imgclip{0}{fig/images/flux_figure_supp/scene_111_prompt} & \imgclip{0}{fig/images/flux_figure_supp/scene_111_gt} & \imgclip{0}{fig/images/flux_figure_supp/scene_111_diorama}  \\
\imgclip{0}{fig/images/flux_figure_supp/scene_09_prompt}& \imgclip{0}{fig/images/flux_figure_supp/scene_09_gt} & \imgclip{0}{fig/images/flux_figure_supp/scene_09_diorama} & \imgclip{0}{fig/images/flux_figure_supp/scene_120_prompt} & \imgclip{0}{fig/images/flux_figure_supp/scene_120_gt} & \imgclip{0}{fig/images/flux_figure_supp/scene_120_diorama}  \\
\imgclip{0}{fig/images/flux_figure/scene_17_prompt} &  \imgclip{0}{fig/images/flux_figure/scene_17_gt} & \imgclip{0}{fig/images/flux_figure/scene_17_diorama} & \imgclip{0}{fig/images/flux_figure/scene_121_prompt} & \imgclip{0}{fig/images/flux_figure/scene_121_gt} & \imgclip{0}{fig/images/flux_figure/scene_121_diorama}  \\
\end{tabularx}
\caption{
More examples of applying Diorama in a text-to-scene setting. 
}
\vspace{-6pt}
\label{fig:text2scene_supp}
\end{figure*}

\subsection{Multimodal 3D shape retrieval}
\label{sec:supp-retrieval}

For each observed object in the image, we use a two-step process for retrieving matching shapes from our shape database.
We first use a text query to retrieve objects matching the semantic category, and then an image query to re-rank the retrieved candidates.  
We find this two-stage approach helps ensure that retrieved objects are of the correct semantic class, as it is possible for semantically different objects to be geometrically similar (e.g. books vs cardboard box). 

We construct the text query from the template \textit{``a photo of CLASS''} (similar to in detection phase). 
For the image query, we use object crop extracted from the original image and mask out background and occluders using bounding box $b_i$ and mask $m_i$. 
We also carefully separate retrieval of supporting and non-supporting objects according to the supporting hierarchy in the scene graph since supporting objects need to be pre-processed to represent only one object entity (not containing smaller sub-objects).

\subsection{Zero-shot object pose estimation.}
\label{sec:supp-pose-estimation}

We pre-compute $T$ multiview renderings and depth maps for each retrieved CAD model $s_i$.
Each query object crop $I^{o_i}$ and set of multiview images $\{I^{s_{i,j}}\}$ is encoded into normalized patch features $\mathcal{F}(I)$ using DinoV2.
For each patch embedding from the query image, we construct a semantic correspondence to a patch embedding from each rendering view with the minimal cyclical distance~\cite{goodwin2022zero, nguyen2024gigaPose}.
We compute correspondence score as cosine similarity between patch embeddings.
Specifically, given a pair of query image $I_q$ and reference rendering image $I_r$, we construct a ``cyclical distance'' map and define a similarity score using their corresponding DINOv2 patch features, $f_q$ and $f_r$, and feature masks $m_q$ and $m_r$. 
For each query feature $f_q^i$ at the patch location $i$, we compute its cyclical patch $i'$ in $I_q$ as follows:
\[
\begin{split}
    i' &= \operatorname*{argmax}_{w, m_q^w>0} S(f_q^w, f_r^j) \\
    j &= \operatorname*{argmax}_{k, m_r^k>0} S(f_q^i, f_r^k)
\end{split}
\]
where $S(\cdot, \cdot)$ denote as cosine similarity and a cyclical distance map is constructed as $D$ with $D_i = -||i, i'||_2$. 
In the end, we build the feature correspondences by taking at most K query-reference patch pairs $(i, j) \in \mathcal{N}$ with top-K minimal cyclical distances in $D$. 
To ensure only keeping correspondences with strong similarity, we empirically set similarity score threshold as 0.7. The overall similarity score between $I_q$ and $I_r$ is defined as the average of similarities of feature correspondences:
$$
\text{sim}(I_q, I_r) = \frac{1}{|\mathcal{N}|} \sum_{(i, j)\in \mathcal{N}} S(f_q^i, f_r^j)
$$
The query image is matched to the most similar multiview rendering with the maximal averaged correspondence-wise feature similarity to produce a coarse pose hypothesis.

\subsection{Scene optimization}
\label{sec:supp-optimization}

During scene optimization, we take the retrieved objects, initial object pose estimates, and optimize the placements so that the relationships in the scene graph are respected. 
In addition to support relations described in the scene graph, we further establish an adherence relationship between each object on the floor and potential walls using a heuristic algorithm that judging whether the closest distance from the object's surface center to a candidate wall is within a predefined threshold.
We hence also optimize object position and rotation based on both the support and adherence relationship.
In the main paper, we described the different stages of optimization.  Here we provide more details on how support is enforced while ensuring there is sufficient space, and no inter-penetrations.

\mypara{Support.} A key relationship is that of support, for which we want to ensure that each object is properly placed in contact with their supporting object.  To do so, for each object in our database, we precompute candidate \emph{support surfaces} by identifying roughly planar surfaces on the object. 
Based on estimated initial object poses and the scene graph $\mathbf{G}$ describing object relationships, for each a pair of interacting objects, we compute a pair of contact surface from the supported object and support surface from the supporting object. The support surface is determined by selecting the surface with the minimal distance to the center of the supported object. The contact surface is selected to be the one having the closest direction with the support direction.

\mypara{Space.} Particularly for the \textit{Space} stage, we define a \emph{supporting volume} for each object by extruding its identified support surface to an extent hitting another surface in the vertical direction. An object is properly supported only if it does not exceed the bounds of its corresponding supporting volume. We formulate the term $e_{\text{vol}}$ as the sum of distance from the corners of the contact surface to the sides of the supporting volume and the vertical distance between the centers of the object bounding box and the supporting volume.

\mypara{Optimization.} In each optimization stage, we use a separate SGD optimizer with initial learning rate 0.01 and momentum 0.9 for corresponding pose parameters, except for the \textit{Space} stage where we set initial learning 0.001 for the scale parameter. We also decay the learning rate by 0.1 every 50 steps. We run 200 steps in total in each optimization stage. We describe the objective function of each optimization stage as below, including weight hyperparameters:
\begin{align}
    \text{Stage 1: }& e_1 = 3 \cdot e_{\text{align}} + e_{\text{sem}} \\
    \text{Stage 2: }& e_2 = 5 \cdot e_{\text{place}} + e_{\text{rel}} \\
    \text{Stage 3: }& e_3 = e_{\text{vol}} \\
    \text{Stage 4: }& e_4 = 5 \cdot e_{\text{place}} + e_{\text{col}} 
\end{align}

\section{Additional experiments}
\label{sec:supp-experiments}

\mypara{Implementation details.}
We render SSDB scene images of size 1008$\times$784. Considering computation efficiency and good coarse pose selection, we render 180 gray-color multiviews of size 224$\times$224 for each 3D shape from predefined camera viewpoints to focus on geometry-wise semantic similarity and leave out effect of texture. For zero-shot pose estimation, we use the fine-tuned ViT-L of DinoV2~\cite{oquab2023dinov2} following~\cite{nguyen2024gigaPose} to embed $14\times14$ image patches. We run experiments on one Nvidia RTX 4090 GPU.

\subsection{Details of comparing against ACDC}
\label{sec:detail-compare-acdc}
We run ACDC on all SSDB images with the default configuration. For both ACDC and Diorama, we feed ground-truth 2D object bounding boxes and segmentation masks into systems to avoid 2D perception errors for analysis.
Both systems retrieve object from out-of-distribution 3D shape collections. Since ACDC is developed upon the OmniGibson simulation platform, it retrieves from the built-in set of approximately 8,800 OmniGibson CAD objects for convenient deployment in the physics engine. For Diorama, since we do not specify certain CAD file formats to be compatible with a physics engine, we are able to include more 3D shapes for retrieval from different sources. In particular, we compose a set of 25K 3D shapes for Diorama to retrieve from. 
For ACDC, overall runtime is dominated by network API calls to the LLM (GPT4o). ACDC calls GPT4o three times per detected object, while Diorama calls GPT4o twice in total, irrespective of the number of objects.
For the user study, we randomly sample 48 images and corresponding results to ask the participants to assess the quality of single-view 3D scene modeling in terms of object matching and overall scene quality. For object matching, we consider both semantic correctness and geometric similarity. For overall scene quality, we consider the accuracy of architecture reconstruction and object arrangement, and physical plausibility of the whole scene. The question order is randomly shuffled to the participant.

\subsection{Architecture reconstruction evaluation}
\label{sec:supp-expr-architecture}

We compare our proposed PlainRecon against a recent method for obtaining 3D room layout via render-and-compare (RaC)~\cite{stekovic2020general}. RaC is a common architecture reconstruction baseline, and though follow-ups exist they introduce marginal improvements while having less reliable or no available public implementation~\cite{Yang_2022_WACV, rozumnyi2023estimating}.
For a fair comparison, we provide RaC with inputs from more modern backbones compared to the ones used in the original implementation. 
We use DepthAnythingV2 or Metric3D depth and PlaneRecTR~\cite{shi2023planerectr} planar segments.

As ACDC outputs plane parameters and masks but does not output actual mesh planes, we perform a simple extraction procedure. First, we back-project the depth used by ACDC to the point cloud. Then, similarly to our method, we run RANSAC for each architectural element based on image segmentation masks to fit the plane, resulting in selecting inlier points that correspond to the largest planar region of the point cloud. We obtain the oriented bounding box from the inliers and extract mesh from it. Finally, we align the normals of the architectural elements with the normals ACDC used to optimize object placement.

\subsection{Object alignment on estimated depth}
\label{sec:supp-expr-alignment-est-depth}

\cref{tab:wss_pose_predd} presents the comparison of different zero-shot pose estimation methods on the 9D CAD alignment task given predicted depth by Metric3D. The results align with our observation under the ground-truth depth setting.

\subsection{Performance for different object groups}
\label{sec:supp-expr-obj-groups}

We also analyze post-optimized object pose according to different object groups each object belongs to in~\cref{tab:wss_groups}. Since we aim for an open-world system that generalizes to long-tail categories in real life, we divide objects into three subgroups: household objects/common furniture, occluded/complete objects, and supported/supporting objects, rather than a coarse set of preselected categories as in prior work~\cite{gumeli2022roca, gao2023diffcad}.
We find that the performance is reduced for dominant household items, occluded objects, and supporting objects.

\subsection{Multiple retrievals}
\label{sec:supp-expr-multiple-retrievals}

In~\cref{tab:wss_hypo}, we investigate the benefits of having more retrieved 3D shapes for correspondence computation and coarse pose proposal. It turns out that alignment accuracy increases with potentially more different pose initialization.

\subsection{Different optimization strategy}
\label{sec:supp-expr-optimization}
We investigate the benefits of using a stage-wise 
optimization procedure rather than a more common all-in-one strategy where all terms are accumulated for optimizing simultaneously in \cref{tab:wss_lo_add}. It shows that we obtain significant gains by decomposing the entire optimization task into separate stages.

\subsection{Quantitative results on ScanNet}
\label{sec:supp-scannet}

\cref{tab:s2c_results} shows quantitative comparison between ROCA, DiffCAD and ours on the proposed evaluation set. 
Following prior work~\cite{gumeli2022roca, sparc, gao2023diffcad}, we report the object alignment accuracy where a CAD model is considered correctly aligned if the translation error $\leq 20$cm, the geodesic rotation error $\leq 20^{\circ}$, and the scale ratio $\leq 20\%$. 
We note that ROCA is trained end-to-end using imperfect Scan2CAD annotations and DiffCAD is trained on the synthetic data per category. Both ROCA and DiffCAD cannot generalize to unseen objects during training (indicated as gray-colored numbers). Our zero-shot method achieves competitive performance against DiffCAD that further exhibits performance degradation under the probabilistic setting due to the partial observations of commonly occluded objects.
\cref{fig:s2c_eval_diff} visually shows differences between ground-truth evaluation sets used in DiffCAD, ROCA and ours.

\section{Qualitative examples}
\label{sec:supp-examples}

We provide additional examples of generated scenes in \Cref{fig:main_results_supp,fig:wss_results_supp,fig:real_supp,fig:text2scene_supp}.
In \Cref{fig:wss_results_supp}, we provide plausible arrangements based on renders from SSDB, as well as the associated predicted scene-graphs. With Diorama, we can produce alternative arrangements that use different objects, while respecting the spatial relationships of the original image (e.g. picture on the wall, monitor on the desk).

We further showcase arrangements from real-world images (\Cref{fig:real_supp}) and images generated via text-to-scene (\Cref{fig:text2scene_supp}).

\end{document}